\documentclass[11pt]{article}
\usepackage[margin=1in]{geometry}
\usepackage{times}
\usepackage{natbib}
\usepackage{iftex}
\ifPDFTeX\else
  \usepackage{fontspec}
\fi

\usepackage{amsmath,amsfonts,bm}

\def\eqref#1{equation~\ref{#1}}

\def\1{\bm{1}}

\DeclareMathAlphabet{\mathsfit}{\encodingdefault}{\sfdefault}{m}{sl}
\SetMathAlphabet{\mathsfit}{bold}{\encodingdefault}{\sfdefault}{bx}{n}

\usepackage{url}
\usepackage{graphicx}
\usepackage{booktabs}
\usepackage{tabularx}
\usepackage{array}
\usepackage[table]{xcolor}
\usepackage{placeins,float,needspace}
\usepackage{wrapfig}
\usepackage[bottom]{footmisc}
\usepackage{hyperref}
\hypersetup{colorlinks=true,linkcolor=blue,citecolor=blue,urlcolor=blue}
\definecolor{MLLMBaselineBg}{HTML}{E8EEF5}
\definecolor{AgentBaselineBg}{HTML}{F5EEDB}
\definecolor{AdSeekGroupBg}{HTML}{E3F2ED}
\definecolor{GainGreen}{HTML}{18794E}
\newsavebox{\MDTableBox}
\newcommand{\MDTableFit}[1]{\sbox{\MDTableBox}{#1}%
    \ifdim\wd\MDTableBox>\linewidth
        \resizebox{\linewidth}{!}{\usebox{\MDTableBox}}%
    \else\usebox{\MDTableBox}\fi}

\title{	Beyond Single Videos: Benchmarking and Active Evidence Seeking for E-Commerce Cross-Video Reasoning}
\author{%
Jinghan Zhao \quad Yiman Hu \quad Liang Wu \quad Jian Xu \quad Bo Zheng\\[0.5em]
\textit{Alibaba Group}%
}
\date{}
\begin{document}
\maketitle

\begin{abstract}
E-commerce videos are information-dense and frequently compared in practice, as consumers compare products and merchants evaluate marketing tactics, making cross-video analysis highly valuable for real-world business analytics.
Yet, existing multimodal models primarily focus on single video understanding, with limited ability to perform the cross-video comparisons needed for e-commerce video analysis.
To bridge this gap, we introduce \textbf{AdsCVR}, the first e-commerce cross-video reasoning benchmark. After filtering, AdsCVR contains 2,483 videos and 6,110 QA pairs across six reasoning dimensions.
Extending reasoning across multiple videos fundamentally magnifies the challenge of pinpointing fine-grained evidence amid massive redundant frames, especially in e-commerce videos where visual details, speech, and on-screen text are combined into densely packed content. We propose \textbf{AdSeek}, an agentic framework that dynamically allocates visual and audio tools during multi-turn exploration, replacing static uniform sampling with active evidence grounding.
While reinforcement learning (RL) can optimize sequential tool use, sparse outcome rewards struggle with long-horizon credit assignment, providing limited guidance on where a trajectory goes wrong or how to correct it.
To address this, we propose an \textbf{offline trajectory rectification mechanism} that identifies and corrects reasoning errors and missing multimodal evidence in RL-generated trajectories. It then uses them for SFT to strengthen the model’s evidence gathering and reasoning abilities, thereby mitigating biases learned during RL.
This mechanism anchors our \textbf{rectified bootstrapping pipeline}: an initial RL phase exposes reasoning bottlenecks, the subsequent SFT phase uses rectified data to calibrate model's thinking and action biases, and a final RL phase achieves optimal convergence.
AdSeek achieves 74.30\% accuracy on the AdsCVR test split, improving upon its backbone (Qwen3-VL-8B-Instruct) by 27.90\%. It also generalizes well to the open-domain CrossVid benchmark, validating its effectiveness in active evidence gathering.

\end{abstract}

\FloatBarrier
\section{Introduction}

Many similar products on e-commerce platforms are promoted through different videos, presenting different marketing strategies. These videos often pack product demonstrations, selling points, and promotional messages into just tens of seconds through visuals, speech, and on-screen text. Consumers and merchants often need to compare this dense multimodal information across videos to inform purchasing and marketing decisions. However, existing advertising benchmarks such as MM-AU, AdsQA, and E-VAds \citep{bose2023mmau,long2025adsqa,liu2026evads} predominantly focus on single-video understanding and do not adequately capture this practical need or the challenges of identifying key differences across videos.


Therefore, we propose \textbf{AdsCVR}, the first e-commerce cross-video reasoning benchmark, evaluating models’ ability to understand complex e-commerce videos and draw effective comparisons across multiple videos.
Unlike general-domain CVR benchmarks such as CrossVid and MVU-Eval \citep{crossvid,peng2025mvueval}, AdsCVR focuses on commercial reasoning over advertisements, where decisive evidence is scattered across videos, non-contiguous temporal intervals, and visual and audio modalities amid distracting dense content.


To improve models' ability to understand and compare e-commerce videos, we propose \textbf{AdSeek}, an agentic framework for multi-turn active evidence acquisition. Unlike static uniform sampling \citep{wu2024longvideobench}, AdSeek dynamically selects which videos, temporal segments, modalities, and visual regions to inspect based on the comparison objective. It filters out irrelevant content and aligns evidence across videos before generating an answer.

\begin{figure}[t]
\centering
\includegraphics[page=1,width=0.88\linewidth]{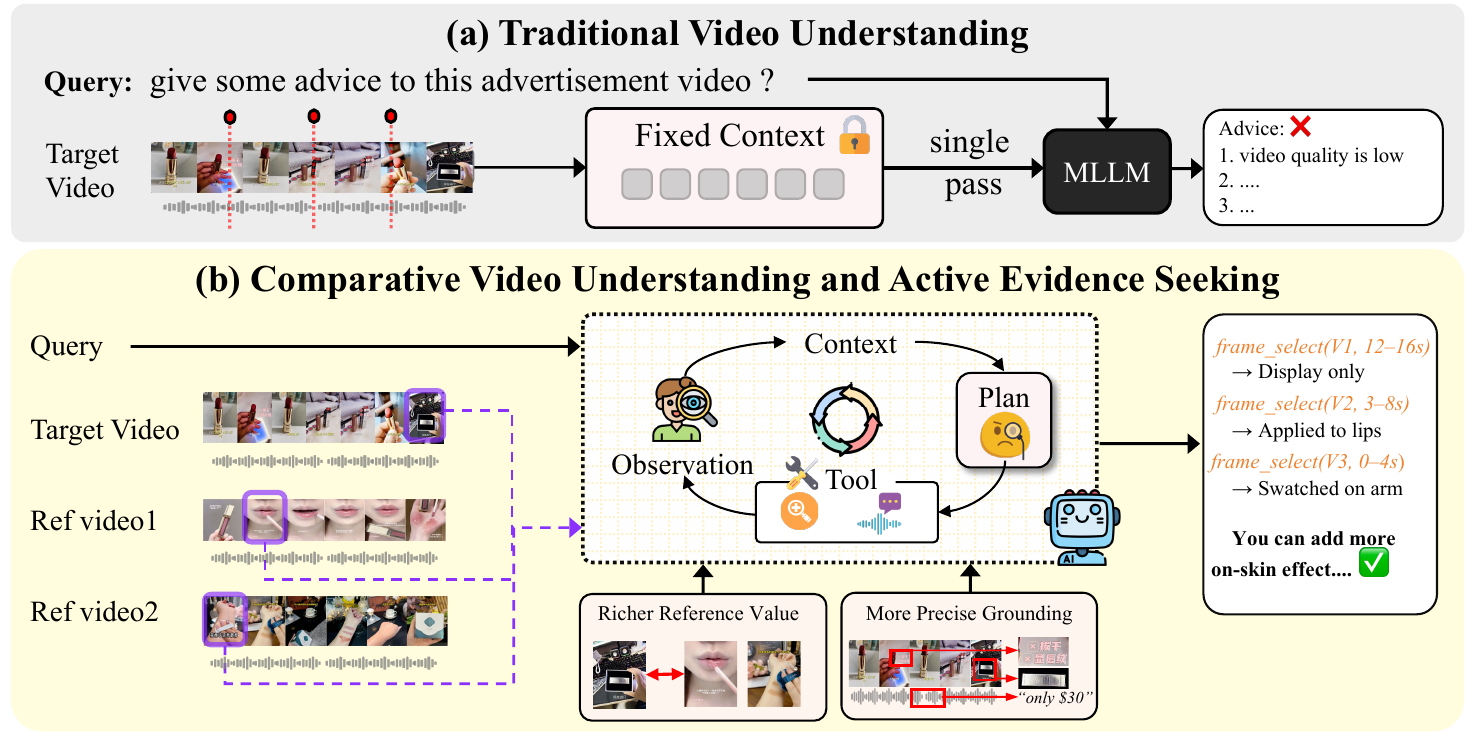}
\caption{\textbf{Single-video understanding versus active cross-video understanding.} (a) Fixed-context, single-pass analysis may produce generic advice. (b) AdSeek leverages richer cross-video references and produce more precise evidence grounding for more specific and actionable advice.}
\label{fig:concept}
\end{figure}

Following tool-augmented language modeling \citep{yao2022react,schick2023toolformer,yang2023gpt4tools,hao2023toolkengpt}, we build a multi-turn interaction environment for AdSeek and train its evidence-search policy through reinforcement learning (RL) \citep{qian2025toolrl}. During RL training process, sparse outcome rewards struggle with long-horizon credit assignment \citep{arjona2019rudder}, failing to provide explicit feedback for intermediate tool calls or reasoning decisions. So, error patterns that emerge during exploration may persist even with continued RL training.

To address this challenge, we introduce an \textbf{offline trajectory rectification mechanism}
. It uses a teacher model to identify intermediate deviations that remain undetected by outcome-based RL feedback and rectify them by supplying missing evidence and correcting faulty reasoning. The corrected trajectories directly supervise intermediate evidence acquisition and reasoning, helping the policy model internalize these corrections via SFT. Building on this, our \textbf{Rectified Bootstrapping Pipeline (RBP)} uses initial RL exploration to expose bottlenecks, offline trajectory rectification to correct persistent biases, and subsequent RL to further refine the policy through online exploration.

Our contributions include:
\begin{itemize}
\item We introduce \textbf{AdsCVR}, an e-commerce cross-video reasoning benchmark comprising 2,483 videos and 6,110 QA pairs spanning 86 product categories (see Table~\ref{tab:benchmark_comparison} for a comparison). It evaluates models’ understanding of complex commercial analysis and their ability to identify fine-grained differences across information-dense videos.
\item We propose \textbf{AdSeek}, an agentic framework that actively acquires and aligns multimodal evidence through multi-turn reasoning and tool use, adapting its search across videos, temporal segments, modalities, and visual regions to each comparison objective.
\item We develop the \textbf{Rectified Bootstrapping Pipeline (RBP)}, which combines online RL exploration with offline trajectory rectification. By correcting biased trajectories and enabling further exploration, this training paradigm strengthens the model's ability to acquire multimodal evidence.
\item AdSeek yields a 74.30\% accuracy on the AdsCVR test split, outperforming the Qwen3-VL-8B-Instruct baseline by an absolute margin of 27.90\%. Beyond e-commerce, it exhibits robust transferability on the open-domain CrossVid dataset, proving its broader capacity for proactive multimodal evidence collection.
\end{itemize}

\section{Related Work}
\subsection*{Video Understanding Benchmarks}
Advances in MLLMs have driven video QA evaluations \citep{lin2023videollava,li2023mvbench} from single-video content and causal reasoning \citep{yu2019activitynetqa,xiao2021nextqa} to long-context understanding \citep{fu2024videomme,zhou2024mlvu,wu2024longvideobench,cheng2025videoholmes} and egocentric tasks \citep{huang2024egoexolearn}. Beyond single videos, benchmarks like CrossVid \citep{crossvid}, MVU-Eval \citep{peng2025mvueval}, and CVBench \citep{zhu2025cvbench} evaluate joint understanding of video collections, requiring models to aggregate distributed evidence and compare entities across videos.

\subsection*{E-commerce Advertisement Video Understanding}
Advertisement understanding requires connecting audiovisual content with persuasive intent. Early works focused on message symbolism \citep{hussain2017automatic}, temporal segmentation, and classification \citep{wang2021tencentavs,bose2023mmau}. Recent QA benchmarks further evaluate visual reasoning in fast-paced ads \citep{zhang2025videoads}, core messages and target audiences \citep{long2025adsqa}, and conversion-oriented e-commerce elements \citep{liu2026evads}. Crucially, these benchmarks evaluate videos individually. In contrast, AdsCVR connects advertisement understanding with cross-video reasoning, requiring models to align cross-ad evidence to compare product presentations.

\begin{table}[H]
\centering
\caption{\textbf{Comparison of video understanding benchmarks.} Temporal and Intent denote explicit temporal and intent understanding coverage ($\checkmark$); ---: not explicitly targeted. MCQ: multiple-choice QA; Open: open-ended QA; Seg.: segmentation; Cls.: classification.}
\label{tab:benchmark_comparison}
\begingroup
\small
\setlength{\tabcolsep}{1.5pt}
\renewcommand{\arraystretch}{1.05}
\begin{tabularx}{\linewidth}{@{}lcrrc>{\centering\arraybackslash}p{0.68in}cc@{}}
\toprule
Benchmark & Domain & \#Videos & \#QA & Video Count & Task format & Temporal & Intent \\
\midrule
MVU-Eval \citep{peng2025mvueval} & General & 4,959 & 1,824 & Multiple & MCQ & $\checkmark$ & --- \\
CrossVid \citep{crossvid} & General & 5,331 & 9,015 & Multiple & MCQ / Open & $\checkmark$ & --- \\
\midrule
Tencent-AVS \citep{wang2021tencentavs} & Ads & 12,000 & N/A & Single & Seg.\ + Cls. & $\checkmark$ & --- \\
MM-AU \citep{bose2023mmau} & Ads & 8,399 & N/A & Single & Cls. & $\checkmark$ & $\checkmark$ \\
VideoAds \citep{zhang2025videoads} & Ads & 200 & 1,100 & Single & MCQ & $\checkmark$ & $\checkmark$ \\
AdsQA \citep{long2025adsqa} & Ads & 1,544 & 7,859 & Single & Open & --- & $\checkmark$ \\
E-VAds \citep{liu2026evads} & E-commerce & 3,961 & 19,785 & Single & Open & --- & $\checkmark$ \\
\midrule
\rowcolor{AdSeekGroupBg}
\textbf{AdsCVR} & E-commerce & \textbf{2,483} & \textbf{6,110} & Multiple & MCQ & $\checkmark$ & $\checkmark$ \\
\bottomrule
\end{tabularx}
\endgroup
\end{table}

\subsection*{MLLMs for Video Understanding}
Video MLLMs acquire temporal reasoning by training visual encoders and language models on multimodal instructions \citep{lin2023videollava,li2023mvbench,zhang2024llavavideo,cheng2024videollama2,wang2024internvideo2}. However, early systems rely on uniform frame sampling, risking the omission of brief, decisive events. To address this, recent agentic and tool-using approaches make perception adaptive, iteratively retrieving or selecting keyframes based on query relevance and reasoning feedback \citep{wang2024videoagent,jeoung2024avua,wang2025videotree,he2025framethinker,zhang2026eva}. AgentCVR extends this active acquisition to cross-video reasoning via script-simulated RL \citep{qiu2026agentcvr}. Together, these approaches shift video understanding from fixed frame sampling toward adaptive evidence acquisition guided by the question and intermediate reasoning.

\section{The AdsCVR Benchmark}
\label{sec:adscvr-benchmark}

\begin{figure}[!b]
\centering
\includegraphics[page=1,width=1.0\linewidth]{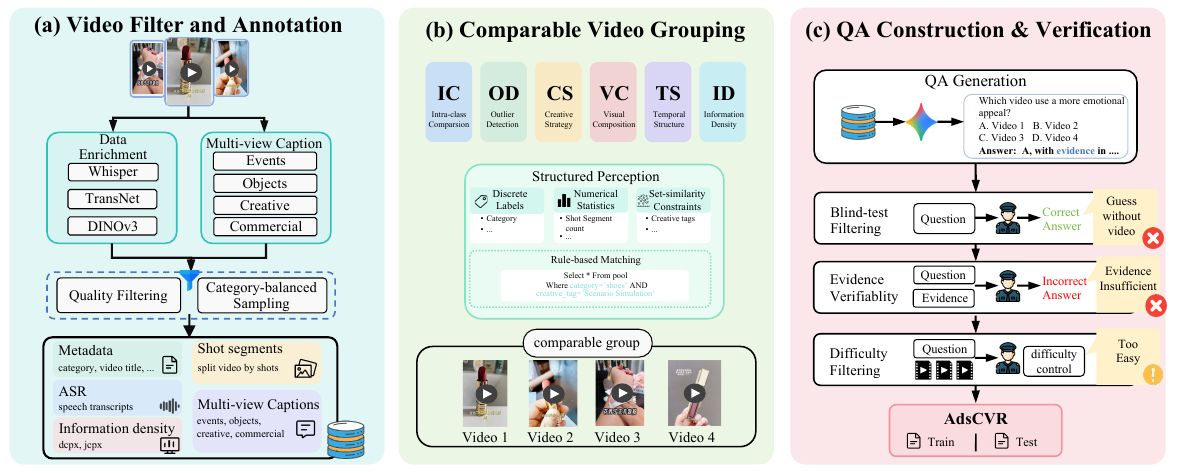}
\caption{\textbf{AdsCVR construction pipeline.} (a) \textbf{Video Annotation:} Raw videos undergo multimodal feature extraction and multidimensional captioning, followed by quality and category-balanced filtering. (b) \textbf{Comparable Grouping:} Structured rules guide the assembly of 2--4 videos into comparable groups across six reasoning dimensions. (c) \textbf{QA Generation \& Filtering:} LLMs generate multiple-choice questions for each group. A rigorous multi-stage filtering pipeline (blind-test, verifiability, and difficulty) yields 6,110 high-quality QA pairs (3,110 train, 3,000 test).}
\label{fig:data_pipeline}
\end{figure}
We introduce the AdsCVR benchmark to evaluate fine-grained e-commerce video understanding. We define a \emph{comparable video group} as a set of videos sharing contextual dimensions (e.g., narrative subject or commercial intent) to expose subtle distinctions in advertising strategies. Figure~\ref{fig:data_pipeline} illustrates our three-stage construction pipeline.


\subsection{Video Filtering and Multi-View Annotation}
\label{sec:video_filtering}
\begin{figure*}[!t]
\centering
\includegraphics[width=0.8\textwidth]{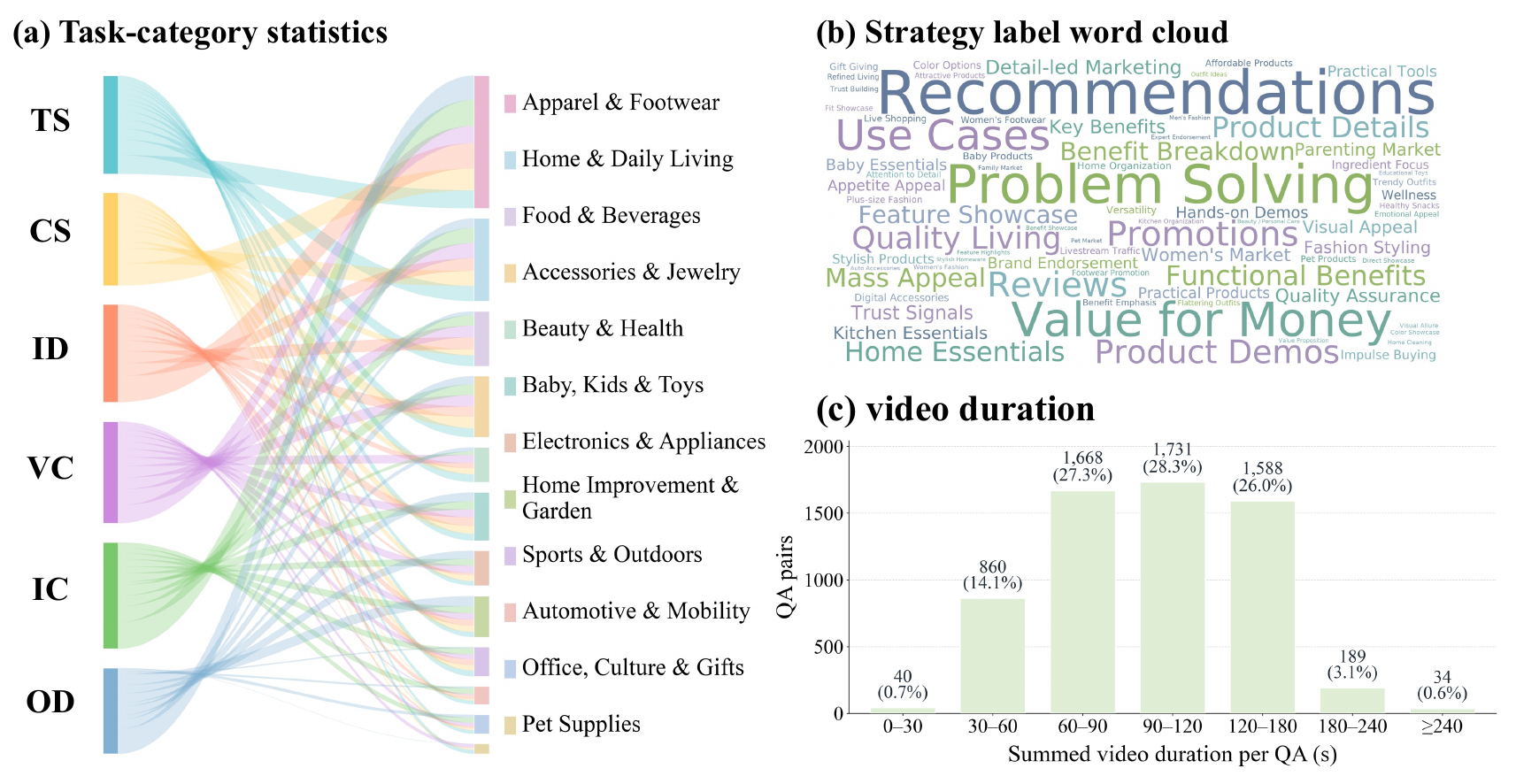}
\caption{\textbf{AdsCVR dataset statistics.} (a) Associations between the six evaluation dimensions and 12 representative product categories, with flow width indicating video count. (b) Commercial-strategy word cloud. (c) Total video duration per QA pair across 6,110 examples.}
\label{fig:adscvr_statistics}
\vspace{-2pt}
\end{figure*}

\paragraph{Video filtering.}
To construct a diverse benchmark of manageable size, we first preprocess the videos to support subsequent filtering:
We extract video metadata, speech transcripts (Whisper-large-v3), and shot boundaries (TransNetV2). To filter out outliers, we estimate temporal and spatial complexities using DINOv3 embeddings and JPEG compression (see Appendix~\ref{app:video_complexity} for metric details). Based on these measures, we discard videos with extreme durations or low visual variation. Finally, we balance the dataset by dropping long-tail categories (fewer than ten videos) and downsampling dominant ones, yielding 2,483 videos across 86 product categories (Figure~\ref{fig:adscvr_statistics}a).

\paragraph{Multi-view annotation.}
To go beyond sparse metadata and capture fine-grained semantics, we prompt Gemini to generate dense, multidimensional captions from four expert perspectives: \emph{events} (actions and demonstrations), \emph{objects} (products and scenes), \emph{creative structure} (narratives and filming techniques), and \emph{commercial logic} (audiences and advertising strategies; see Figure~\ref{fig:adscvr_statistics}b). The full multi-view annotation prompt is provided in Appendix~\ref{app:multiview_prompt}. These annotations provide the semantic foundation for the subsequent video grouping and QA generation.

\subsection{Video Grouping}
\label{sec:video_grouping}
We form comparable groups of two to four videos using dimension-specific sampling rules (Table~\ref{tab:video_grouping}), covering visual execution and commercial intent.

\begin{table}[H]
\centering
\caption{AdsCVR comparison dimensions and video-grouping rules.}
\label{tab:video_grouping}
\begingroup
\fontsize{8.5}{9.2}\selectfont
\setlength{\tabcolsep}{4pt}
\renewcommand{\arraystretch}{1.0}
\begin{tabularx}{\linewidth}{@{}>{\raggedright\arraybackslash}p{1.75in}>{\raggedright\arraybackslash}X@{}}
\toprule
Dimension & Comparison objective and sampling rule \\
\midrule
\textbf{Intra-class Comparison (IC)} & Fine-grained discrimination: sample unique video combinations within the same first-level category. \\
\textbf{Outlier Detection (OD)} & Anomaly identification: combine $N-1$ reference videos with one outlier from a different second-level or first-level category, with a $50\%$/$50\%$ allocation. \\
\textbf{Creative Strategy (CS)} & Commercial intent: include at least two distinct commercial-strategy buckets per group. \\
\textbf{Visual Composition (VC)} & Visual presentation: require creative-tag Jaccard similarity $<0.5$ between candidate and seed videos. \\
\textbf{Temporal Structure (TS)} & Narrative progression: use segment count as a proxy and span at least two ranges (1--3, 4--6, $\geq 7$ segments) per group. \\
\textbf{Information Density (ID)} & Visual pacing: sample intra-category videos whose absolute shot counts differ by $\geq 3$, independent of duration. \\
\bottomrule
\end{tabularx}
\endgroup
\end{table}

\subsection{QA Generation and Verification}
\label{sec:qa_generation}
To construct informative and challenging multiple-choice QA pairs, we employ prompt-based generation followed by a rigorous multistage filtering pipeline. First, for \textbf{QA generation}, Gemini-3.5-Flash generates a single-choice question, options, the correct answer, and supporting textual clues based on aggregated video metadata, multidimensional captions, and dimension-specific grouping rationales. Next, we apply \textbf{blind-test filtering} to mitigate language-prior leakage, discarding any QA pair that Qwen3VL-8B answers correctly in three consecutive attempts using exclusively the text question and options (without video inputs). To ensure \textbf{answer verifiability} and grounding in video evidence, we then require GPT-5.5 to correctly predict the reference answer relying solely on the generated textual clues. Finally, for \textbf{difficulty filtering}, we evaluate the remaining pairs using Qwen3-VL-8B-Instruct with uniformly sampled frames, randomly discarding 50\% of the questions it answers correctly to calibrate the overall benchmark difficulty.





AdsCVR contains 6,110 QA pairs from 2,483 e-commerce videos across 86 top-level product categories, with QA pairs randomly split into 3,110 training and 3,000 test examples.

\section{AdSeek}

\begin{figure}[t]
\centering
\includegraphics[page=1,width=0.98\linewidth]{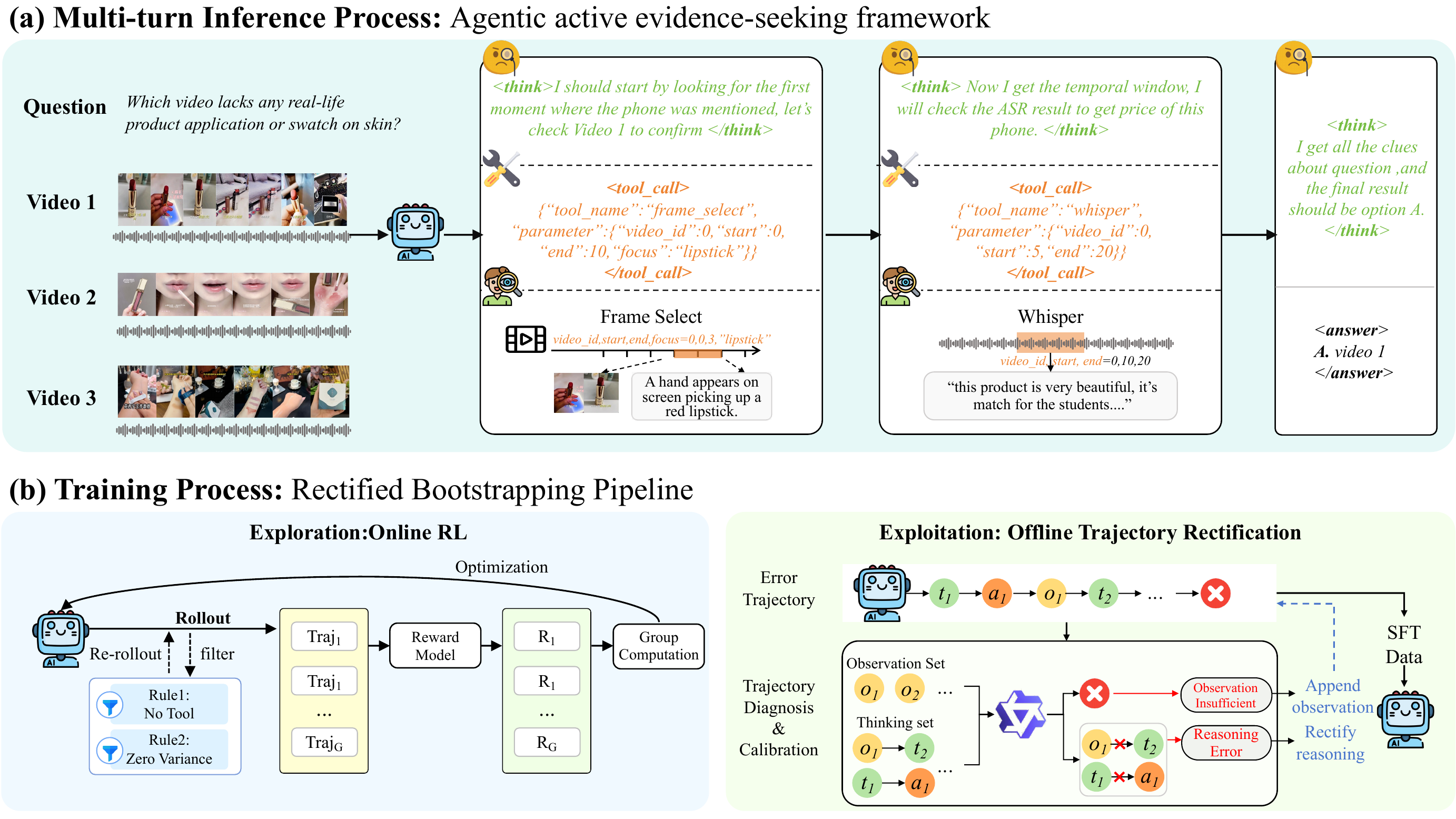}
\caption{\textbf{AdSeek overview.} (a) AdSeek’s multi-turn inference process, where reasoning guides tool use to actively acquire relevant evidence. (b) The two modules of the Rectified Bootstrapping Pipeline: online RL exploration with trajectory filtering, and offline teacher-guided trajectory rectification, which corrects observation and reasoning errors and provides targeted SFT supervision for subsequent exploration.}
\label{fig:method_overview}
\end{figure}

We introduce \textbf{AdSeek}, a framework for active multi-turn evidence acquisition trained through the Rectified Bootstrapping Pipeline, as illustrated in Figure~\ref{fig:method_overview}.

\subsection{Task Definition}
\label{sec:task_definition}
Given a query $q$ and a video collection $\mathcal{V}=\{V_j\}_{j=1}^{N}$, the task is to generate an answer $\hat{y}$ by iteratively acquiring cross-video evidence via perception tools.

\textbf{Interaction process.} The policy $\Phi$ initializes with context $h_1=(q,\mathcal{M})$, where $\mathcal{M}$ contains lightweight metadata (e.g., duration, resolution). At round $i$, the context expands to:
\begin{equation}
h_i=(q,\mathcal{M},t_1,a_1,o_1,\ldots,t_{i-1},a_{i-1},o_{i-1}),
\end{equation}
where $t_k$, $a_k$, and $o_k$ denote the reasoning trace, tool call, and returned textual observation at round $k$. Conditioned on $h_i$, $\Phi$ generates $t_i$ followed by either a final answer $\hat{y}$ or a tool call $a_i\in\mathcal{A}$. Executing $a_i$ yields observation $o_i$, updating the context as:
\begin{equation}
h_{i+1}=(h_i,t_i,a_i,o_i).
\end{equation}

\textbf{Perception tools.} The action space comprises visual and audio tools:
\begin{equation}
\begin{aligned}
\mathcal{A}
&=\mathcal{A}_{\mathrm{vis}}\cup\mathcal{A}_{\mathrm{aud}},\\
\mathcal{A}_{\mathrm{vis}}
&=\{\texttt{frame\_select}(j,s,e,f)\},\\
\mathcal{A}_{\mathrm{aud}}
&=\{\texttt{audio\_select}(j,s,e)\},
\end{aligned}
\end{equation}
where $j\in\{1,\ldots,N\}$ identifies the video and $0\leq s<e\leq D_j$ specifies a temporal interval. The prompt $f\in\mathcal{F}$ defines specific visual details to inspect. The visual and audio tools return a focused textual description and a speech transcript of the segment, respectively.

\textbf{Trajectory.} This process repeats until $\Phi$ produces an answer or hits the interaction limit, yielding a completed $n$-round trajectory:
\begin{equation}
\tau=(q,\mathcal{M},t_1,a_1,o_1,\ldots,t_{n-1},a_{n-1},o_{n-1},t_n,\hat{y}).
\end{equation}

\subsection{Exploration: Online Reinforcement Learning (RL)}
\label{sec:rl}

\subsubsection{Policy Optimization}
\label{sec:policy_optimization}
We optimize the policy using Group Relative Policy Optimization (GRPO) \citep{shao2024deepseekmath}, implemented with ROLL \citep{wang2025roll}. For each input $x$, the policy samples $G$ trajectories $\{\tau_i\}_{i=1}^{G}$. Each trajectory receives a reward combining final-answer correctness (w.r.t.\ the reference $y^\star$) and action-format compliance:
\begin{equation}
R(\tau_i)=R_{\mathrm{acc}}(\hat{y}_i,y^\star)+0.1R_{\mathrm{fmt}}(\tau_i).
\label{eq:trajectory_reward}
\end{equation}
We compute the group-normalized advantage as $\widehat{A}_i = (R(\tau_i)-\mu_R)/(\sigma_R+\epsilon_A)$, where $\mu_R$ and $\sigma_R$ are the group reward mean and standard deviation. For a generated token $z_{i,k}$ with preceding context $c_{i,k}$, we define the probability ratio $r_{i,k}(\theta) = \pi_\theta(z_{i,k}\mid c_{i,k}) / \pi_{\theta_{\mathrm{old}}}(z_{i,k}\mid c_{i,k})$. The token-level GRPO objective is then:
\begin{equation}
\begin{aligned}
\mathcal{L}_{\mathrm{GRPO}}(\theta)
=-\mathbb{E}_{i,k}\Big[
&\min\big(
r_{i,k}(\theta)\widehat{A}_i,
\operatorname{clip}(r_{i,k}(\theta),1-\epsilon,1+\epsilon)\widehat{A}_i
\big)\\
&-\beta D_{\mathrm{KL}}\big(
\pi_\theta(\cdot\mid c_{i,k})
\| \pi_{\mathrm{ref}}(\cdot\mid c_{i,k})
\big)
\Big],
\end{aligned}
\end{equation}
where $\epsilon$ controls policy-ratio clipping and $\beta$ weights the KL penalty.

\subsubsection{Rollout Filtering and Resampling}
\label{sec:rollout_filtering}
We apply two filtering rules to encourage evidence acquisition and retain informative reward comparisons:
(1)~\textbf{No-tool trajectories.} We discard trajectories with no effective tool calls, $N_{\mathrm{action}}=0$. Since video content is accessible only through perception tools, these trajectories provide no evidence of video inspection. (2)~\textbf{Zero-variance groups.} We discard groups $\mathcal{G}$ with $\operatorname{Var}(\{R(\tau):\tau\in\mathcal{G}\})=0$. Such groups yield zero normalized advantages and therefore no reward-driven policy-gradient signal. They can arise when all trajectories receive the same correctness and format rewards.
After filtering, we collect additional rollouts to replenish the training batch, allowing up to three (\texttt{MAX\_TRY}=3) resampling attempts. Implementation details are provided in Appendix~\ref{app:implementation}.

\subsection{Exploitation: Offline Trajectory Rectification}
\label{sec:trajectory_repair}
\label{sec:sft}
Outcome rewards provide limited guidance on intermediate failures \citep{lightman2023letsverify}. Building on feedback-guided refinement \citep{madaan2023selfrefine,shinn2023reflexion,havrilla2024glore}, we introduce offline trajectory rectification to diagnose and correct deviations in the policy's exploration. This diagnosis targets two core issues: evidence sufficiency and reasoning validity.

\paragraph{Candidate selection.}
We select failed training trajectories from the RL-trained policy where a stronger teacher ($\Phi_{\mathrm{teacher}}$) succeeds in the same environment. The teacher's successful trajectory serves as a reference for identifying evidence gaps and guiding corrections.

\paragraph{Observation sufficiency check.}
For a trajectory with observations $\mathcal{O}_{\tau}=\{o_1,\ldots,o_{n-1}\}$, the teacher attempts to answer the query using only $\mathcal{O}_{\tau}$ (without tool calls). An incorrect answer acts as a diagnostic signal indicating a potential evidence gap (e.g., missing videos, intervals, or modalities).

\paragraph{Reasoning validity check.}
The teacher reviews each reasoning step against three criteria: (1)~\textbf{Factual consistency} (claims must be grounded in observations, avoiding hallucinations), (2)~\textbf{Reasoning completeness} (conclusions require sufficient evidence support), and (3)~\textbf{Historical consistency} (alignment with prior rounds). Steps are flagged if any criterion's risk score (1--5) exceeds 3.

\paragraph{Targeted trajectory rectification.}
We apply corrections based on the diagnosis: (1)~\textbf{Evidence omission:} We append missing observations from the teacher's reference trajectory to the policy's context, prompting $\Phi$ to regenerate intermediate reasoning. (2)~\textbf{Reasoning deviation:} The teacher provides corrective comments on flagged steps to guide $\Phi$'s reasoning regeneration. By targeting specific failures within the student's own context (rather than replacing it entirely with the teacher's rollout), we produce tailored corrective demonstrations.

\subsubsection{Supervised Policy Optimization}
Following rationale bootstrapping \citep{zelikman2022star}, we construct $\mathcal{D}_{\mathrm{SFT}}$ from initially successful and newly rectified trajectories, helping the policy internalize corrected evidence acquisition and reasoning.

For a tokenized trajectory $z_{1:M}$, a mask $m_j\in\{0,1\}$ isolates assistant tokens (reasoning traces, tool calls, and final answers, where $m_j=1$) from the context (queries, metadata, and observations, where $m_j=0$). The objective is:
\begin{equation}
\mathcal{L}_{\mathrm{SFT}}(\theta)
=-\mathbb{E}_{\tau\sim\mathcal{D}_{\mathrm{SFT}}}
\left[
\frac{\sum_{j=1}^{M}m_j\log\pi_\theta(z_j\mid z_{<j})}
{\sum_{j=1}^{M}m_j}
\right].
\end{equation}
The loss is averaged solely over assistant tokens. This applies direct supervision to intermediate steps on fixed trajectories, without invoking perception tools during training.

\subsection{Training Paradigm: Rectified Bootstrapping Pipeline}
\label{sec:training_pipeline}
We integrate these components into the \emph{Rectified Bootstrapping Pipeline} (RBP), a three-stage paradigm: (1)~\textbf{Initial RL} develops evidence acquisition and exposes trajectory errors; (2)~\textbf{Offline Rectification \& SFT} corrects missing evidence and faulty reasoning via teacher guidance, internalized through SFT; (3)~\textbf{Refined RL} resumes online optimization. By complementing sparse outcome feedback with direct supervision of intermediate decisions, RBP strengthens the model's ability to locate, integrate, and compare evidence scattered across videos, temporal segments, and modalities, supporting a deeper understanding of information-dense e-commerce videos.

\FloatBarrier
\section{Experiments}

\subsection*{Training and Evaluation Settings}
Our policy backbone is Qwen3-VL-8B-Instruct \citep{bai2025qwen3vl}, using Qwen3-VL-4B-Instruct and Whisper large-v3 \citep{radford2022whisper} for perception, and Qwen3.8-27B for offline attribution and evaluation. We use 16 PPUs for SFT via LLaMA-Factory \citep{zheng2024llamafactory} and 32 PPUs for RL via ROLL \citep{wang2025roll} (Appendix~\ref{app:implementation}). We evaluate accuracy on the AdsCVR test split (3,000 test / 3,110 train from 6,110 QA pairs) and CrossVid for out-of-domain generalization. Baselines include general MLLMs and AgentCVR \citep{qiu2026agentcvr}\footnote{Reproduced in our own environment using the same infrastructure, based on the code released by the authors.}, a prior cross-video RL framework.
\subsection{In-Domain Results on AdsCVR}
\label{sec:in_domain}

Table~\ref{tab:adscvr_results} highlights the difficulty of AdsCVR: evaluated open-source single-pass models achieve average accuracies of 46.40\%--75.15\%, leaving substantial room for improvement. Trained only on AdsCVR, AdSeek reaches 73.57\%, outperforming its Qwen3-VL-8B backbone by 27.17 percentage points and our AgentCVR reproduction by 2.52 points, while approaching Qwen3.8-27B (75.15\%). Incorporating public out-of-domain training data further improves accuracy to 74.30\%; we analyze mixed-source training in Section~\ref{sec:mixed_training}.
\par
Average accuracy improves across training stages, with the largest stage-to-stage gain occurring after RL resumes (+2.04 points). Relative to Stage 1, the final mixed-source model improves across all six dimensions, most notably Information Density (ID, +5.0 points) and Visual Composition (VC, +4.8 points), where answering requires locating fine-grained evidence in dense visual content or comparing details across shots. Together, these results suggest that AdSeek’s SFT stage preserves overall performance while enabling further improvement during subsequent RL, and that targeted trajectory rectification contributes to higher accuracy on challenging tasks involving dense visual content and cross-shot comparisons.

\begin{table}[!hb]
\centering

\caption{Accuracy (\%) on the AdsCVR test split. All AdSeek variants use Qwen3-VL-8B. Bold indicates our best results. Average denotes the unweighted mean across dimensions.}
\label{tab:adscvr_results}
\begingroup
\small
\setlength{\tabcolsep}{3pt}
\renewcommand{\arraystretch}{1.0}
\MDTableFit{\begin{tabular}{@{}lrrrrrrr@{}}
\toprule
Method & TS & CS & ID & IC & OD & VC & Average \\
\midrule
\rowcolor{MLLMBaselineBg}
\multicolumn{8}{@{}l}{\textit{General-purpose MLLM baselines}} \\
Gemini 3.5 Flash & 79.3 & 75.0 & 76.2 & 75.6 & 78.8 & 81.5 & 77.73 \\
GPT-5.5 & 85.2 & 80.2 & 84.3 & 79.6 & 82.0 & 83.1 & 82.40 \\
Qwen3.8-27B & 79.7 & 74.3 & 74.9 & 77.1 & 72.2 & 72.7 & 75.15 \\
Qwen3.5-4B & 64.8 & 59.0 & 56.6 & 61.0 & 60.6 & 61.9 & 60.65 \\
Qwen3.5-9B & 68.9 & 64.6 & 67.1 & 64.0 & 64.9 & 64.5 & 65.67 \\
\addlinespace[1pt]
\rowcolor{AgentBaselineBg}
\multicolumn{8}{@{}l}{\textit{Prior agentic baseline}} \\
AgentCVR (2026) & 73.7 & 71.1 & 68.7 & 70.3 & 73.6 & 68.9 & 71.05 \\
\addlinespace[1pt]
\rowcolor{AdSeekGroupBg}
\multicolumn{8}{@{}l}{\textit{AdSeek with Qwen3-VL-8B backbone}} \\
Qwen3-VL-8B (backbone) & 52.2 & 46.8 & 41.6 & 43.6 & 48.2 & 46.0 & 46.40 \\
AdSeek (Stage 1) & 74.6 & 72.4 & 65.8 & 72.8 & 72.6 & 69.4 & 71.27 \\
AdSeek (Stage 2) & 75.0 & 67.6 & 68.2 & 73.6 & 74.4 & 70.4 & 71.53 \\
AdSeek (Stage 3, in-domain only) & 75.0 & 74.4 & 70.6 & 72.8 & 74.6 & 74.0 & 73.57 \\
\textbf{AdSeek (Stage 3, mixed)} & \textbf{77.0} & \textbf{73.4} & \textbf{70.8} & \textbf{75.6} & \textbf{74.8} & \textbf{74.2} & \textbf{74.30} \\
\textcolor{GainGreen}{Gain vs. backbone (pp)} & \textcolor{GainGreen}{+24.8} & \textcolor{GainGreen}{+26.6} & \textcolor{GainGreen}{+29.2} & \textcolor{GainGreen}{+32.0} & \textcolor{GainGreen}{+26.6} & \textcolor{GainGreen}{+28.2} & \textcolor{GainGreen}{+27.90} \\
\bottomrule
\end{tabular}}
\endgroup
\end{table}




\subsection{Out-of-Domain Results on CrossVid}
\label{sec:crossvid_transfer}

To assess cross-domain generalization, we evaluate AdSeek on ten tasks from CrossVid, a general-domain cross-video benchmark. As shown in Table~\ref{tab:crossvid_results}, AdSeek achieves an average accuracy of 31.55\%, exceeding its Qwen3-VL-8B backbone by 2.08 percentage points and our AgentCVR reproduction by 3.31 points. It improves over the backbone on six of the ten tasks. These results suggest that AdSeek's evidence-seeking capabilities transfer beyond e-commerce videos.


\begin{table}[!htb]
\centering
\caption{Out-of-domain performance (\%) on CrossVid. Average is the unweighted mean over ten tasks; means and gains are computed before display rounding. Bold marks the highest average.}
\label{tab:crossvid_results}
\begingroup
\small
\setlength{\tabcolsep}{3pt}
\renewcommand{\arraystretch}{1.0}
\MDTableFit{\begin{tabular}{@{}lrrrrrrrrrrr@{}}
\toprule
Method & BU & NC & CC & PEA & PI & FSA & MSR & MOC & PSS & CCQA & Average \\
\midrule
Qwen3-VL-8B & 23.5 & 27.4 & 40.7 & 34.9 & 62.5 & 5.71 & 34.5 & 27.2 & 7.22 & 31.2 & 29.47 \\
AgentCVR (2026) & 27.6 & 37.5 & 48.9 & 30.1 & 50.2 & 11.0 & 22.3 & 28.4 & 8.14 & 18.2 & 28.24 \\
AdSeek (Stage 3, mixed) & 38.9 & 38.7 & 47.7 & 37.5 & 45.4 & 16.1 & 26.4 & 37.8 & 6.61 & 20.4 & \textbf{31.55} \\
\bottomrule
\end{tabular}}
\endgroup
\end{table}

\subsection{Ablations on Training Data and Action Space}
\label{sec:mixed_training}

\paragraph{Training data.}
Table~\ref{tab:training_mixture} shows that AdsCVR-only training outperforms MVU-Eval-only training by 9.50 percentage points (73.57\% vs.\ 64.07\%), supporting the value of domain-specific data for e-commerce video comparison. Our mixed-source dataset, AdSeek-train, supplements AdsCVR with MVU-Eval videos and script-simulated AgentCVR examples, further improving accuracy to 74.30\%. The additional gains are concentrated in IC (+2.8 points) and TS (+2.0 points), while CS decreases by 1.0 point. Thus, mixed-source training provides a modest overall benefit, with effects that vary across comparison dimensions.
\paragraph{Action space.}
Restricting AdSeek to \texttt{frame\_select} or \texttt{audio\_select} reduces average accuracy to 63.97\% and 65.03\%, respectively, corresponding to drops of 10.33 and 9.27 points from the full model. The two modalities exhibit different strengths: visual-only access performs better on IC (63.8\% vs.\ 58.2\%), whereas speech-only access performs better on the other five dimensions. Nevertheless, the full model outperforms both restricted variants across all six dimensions, supporting the benefit of combining visual and speech evidence for cross-video comparison.

\setlength{\intextsep}{4pt}
\begin{table}[!htb]
\centering
\caption{Training-data and action-space ablations on AdsCVR test accuracy (\%). Average denotes the unweighted mean across dimensions. MVU-Eval-only and restricted-action variants are evaluated on the 3,000-question test split, with unparsed answers counted as incorrect.}
\label{tab:training_mixture}
\begingroup
\small
\setlength{\tabcolsep}{3pt}
\renewcommand{\arraystretch}{0.95}
\MDTableFit{\begin{tabular}{@{}lrrrrrrr@{}}
\toprule
Variant & TS & CS & ID & IC & OD & VC & Average \\
\midrule
\multicolumn{8}{@{}l}{\textit{Training data}} \\
In-domain only (AdsCVR) & 75.0 & 74.4 & 70.6 & 72.8 & 74.6 & 74.0 & 73.57 \\
MVU-Eval only & 67.4 & 60.8 & 60.8 & 63.6 & 65.6 & 66.2 & 64.07 \\
Mixed & 77.0 & 73.4 & 70.8 & 75.6 & 74.8 & 74.2 & 74.30 \\
\midrule
\multicolumn{8}{@{}l}{\textit{Action space}} \\
\texttt{frame\_select} only & 66.2 & 63.8 & 61.2 & 63.8 & 65.8 & 63.0 & 63.97 \\
\texttt{audio\_select} only & 67.4 & 65.6 & 65.6 & 58.2 & 66.2 & 67.2 & 65.03 \\
\textbf{AdSeek} & \textbf{77.0} & \textbf{73.4} & \textbf{70.8} & \textbf{75.6} & \textbf{74.8} & \textbf{74.2} & \textbf{74.30} \\
\bottomrule
\end{tabular}}
\endgroup
\end{table}
\setlength{\intextsep}{8pt}

\section{Limitations}
AdSeek involves computational trade-offs associated with multi-turn agentic reasoning. Sequential tool calls and repeated context processing can increase inference latency, while RL rollouts and teacher-guided trajectory rectification introduce additional training overhead. Future work could explore more efficient tool scheduling and lightweight perception models to better balance reasoning quality and computational cost, particularly for latency-sensitive applications.
\section{Conclusion}
We introduced AdsCVR, a comprehensive benchmark for fine-grained cross-video e-commerce comparison, alongside AdSeek, an agentic framework for active evidence acquisition. To optimize this framework, we developed the Rectified Bootstrapping Pipeline (RBP), which seamlessly integrates RL exploration with offline trajectory rectification via SFT. AdSeek achieves 74.30\% accuracy on AdsCVR and demonstrates robust transferability to open-domain multimodal reasoning tasks.

\FloatBarrier
\section*{AI Use Statement}
We used generative AI models as components of the reported research workflow. Gemini models were used to produce multi-view video annotations and candidate AdsCVR questions, answer options, and supporting clues. Qwen3-VL-8B was used for prompt-only and frame-based filtering, GPT-5.5 was used for evidence-based answer verification, and Qwen3.8-27B was used as the teacher for offline trajectory diagnosis and as an evaluation judge where described in the paper. These model outputs were processed by deterministic filters and checked on representative examples by the authors; model-generated outputs were not treated as ground truth without the stated verification steps. The authors reviewed the final manuscript, experimental claims, and released code, and take responsibility for the accuracy, integrity, and originality of all submitted material.

\bibliographystyle{plainnat}
\bibliography{reference}

\begin{thebibliography}{39}
\bibitem[Arjona-Medina et~al.(2019)]{arjona2019rudder}
Jose A. Arjona-Medina, Michael Gillhofer, Michael Widrich, Thomas Unterthiner, Johannes Brandstetter, and Sepp Hochreiter. RUDDER: Return decomposition for delayed rewards. \emph{arXiv preprint arXiv:1806.07857}, version 3, 2019. URL \url{https://arxiv.org/abs/1806.07857}.

\bibitem[Bai et~al.(2025)]{bai2025qwen3vl}
Shuai Bai, Yuxuan Cai, Ruizhe Chen, et~al. Qwen3-VL technical report. \emph{arXiv preprint arXiv:2511.21631}, 2025. URL \url{https://arxiv.org/abs/2511.21631}.

\bibitem[Bose et~al.(2023)]{bose2023mmau}
Digbalay Bose, Rajat Hebbar, Tiantian Feng, Krishna Somandepalli, Anfeng Xu, and Shrikanth Narayanan. MM-AU: Towards multimodal understanding of advertisement videos. \emph{arXiv preprint arXiv:2308.14052}, 2023. URL \url{https://arxiv.org/abs/2308.14052}.

\bibitem[Cheng et~al.(2024)]{cheng2024videollama2}
Zesen Cheng, Sicong Leng, Hang Zhang, Yifei Xin, Xin Li, Guanzheng Chen, Yongxin Zhu, Wenqi Zhang, Ziyang Luo, Deli Zhao, and Lidong Bing. VideoLLaMA 2: Advancing spatiotemporal modeling and audio understanding in video LLMs. \emph{arXiv preprint arXiv:2406.07476}, 2024. URL \url{https://arxiv.org/abs/2406.07476}.

\bibitem[Cheng et~al.(2025)]{cheng2025videoholmes}
Junhao Cheng, Yuying Ge, Teng Wang, Yixiao Ge, Jing Liao, and Ying Shan. Video-Holmes: Can MLLM think like Holmes for complex video reasoning? \emph{arXiv preprint arXiv:2505.21374}, 2025. URL \url{https://arxiv.org/abs/2505.21374}.

\bibitem[Feng et~al.(2025)]{feng2025videor1}
Kaituo Feng, Kaixiong Gong, Bohao Li, Zonghao Guo, Yibing Wang, Tianshuo Peng, Junfei Wu, Xiaoying Zhang, Benyou Wang, and Xiangyu Yue. Video-R1: Reinforcing video reasoning in MLLMs. \emph{arXiv preprint arXiv:2503.21776}, 2025. URL \url{https://arxiv.org/abs/2503.21776}.

\bibitem[Fu et~al.(2024)]{fu2024videomme}
Chaoyou Fu, Yuhan Dai, Yongdong Luo, et~al. Video-MME: The first comprehensive evaluation benchmark of multimodal LLMs for video analysis. \emph{arXiv preprint arXiv:2405.21075}, 2024. URL \url{https://arxiv.org/abs/2405.21075}.

\bibitem[Grauman et~al.(2022)]{grauman2022ego4d}
Kristen Grauman, Andrew Westbury, Eugene Byrne, et~al. Ego4D: Around the world in 3,000 hours of egocentric video. In \emph{Proceedings of the IEEE/CVF Conference on Computer Vision and Pattern Recognition}, pp. 18995--19012, 2022.

\bibitem[Hao et~al.(2023)]{hao2023toolkengpt}
Shibo Hao, Tianyang Liu, Zhen Wang, and Zhiting Hu. ToolkenGPT: Augmenting frozen language models with massive tools via tool embeddings. In \emph{Advances in Neural Information Processing Systems}, 2023. URL \url{https://proceedings.neurips.cc/paper_files/paper/2023/hash/8fd1a81c882cd45f64958da6284f4a3f-Abstract-Conference.html}.

\bibitem[Havrilla et~al.(2024)]{havrilla2024glore}
Alexander Havrilla, Sharath Chandra Raparthy, Christoforos Nalmpantis, Jane Dwivedi-Yu, Maksym Zhuravinskyi, Eric Hambro, and Roberta Raileanu. GLoRe: When, where, and how to improve LLM reasoning via global and local refinements. In \emph{Proceedings of the International Conference on Machine Learning}, 2024. URL \url{https://proceedings.mlr.press/v235/havrilla24a.html}.

\bibitem[He et~al.(2025)]{he2025framethinker}
Zefeng He, Xiaoye Qu, Yafu Li, Siyuan Huang, Daizong Liu, and Yu Cheng. FrameThinker: Learning to think with long videos via multi-turn frame spotlighting. \emph{arXiv preprint arXiv:2509.24304}, 2025. URL \url{https://arxiv.org/abs/2509.24304}.

\bibitem[Huang et~al.(2024)]{huang2024egoexolearn}
Yifei Huang, Guo Chen, Jilan Xu, Mingfang Zhang, Lijin Yang, Baoqi Pei, Hongjie Zhang, Lu Dong, Yali Wang, Limin Wang, and Yu Qiao. EgoExoLearn: A dataset for bridging asynchronous ego- and exo-centric view of procedural activities in real world. In \emph{Proceedings of the IEEE/CVF Conference on Computer Vision and Pattern Recognition}, pp. 22072--22086, 2024.

\bibitem[Hussain et~al.(2017)]{hussain2017automatic}
Zaeem Hussain, Mingda Zhang, Xiaozhong Zhang, et~al. Automatic understanding of image and video advertisements. In \emph{Proceedings of the IEEE Conference on Computer Vision and Pattern Recognition}, pp. 1705--1715, 2017.

\bibitem[Jeoung et~al.(2024)]{jeoung2024avua}
Sullam Jeoung, Goeric Huybrechts, Bhavana Ganesh, Aram Galstyan, and Sravan Bodapati. Adaptive video understanding agents: Enhancing efficiency through dynamic frame sampling and feedback-driven reasoning. \emph{arXiv preprint arXiv:2410.20252}, 2024. URL \url{https://arxiv.org/abs/2410.20252}.

\bibitem[Jang et~al.(2017)]{jang2017tgifqa}
Yunseok Jang, Yale Song, Youngjae Yu, Youngjin Kim, and Gunhee Kim. TGIF-QA: Toward spatio-temporal reasoning in visual question answering. In \emph{Proceedings of the IEEE Conference on Computer Vision and Pattern Recognition}, 2017. URL \url{https://openaccess.thecvf.com/content_cvpr_2017/html/Jang_TGIF-QA_Toward_Spatio-Temporal_CVPR_2017_paper.html}.

\bibitem[Lei et~al.(2018)]{lei2018tvqa}
Jie Lei, Licheng Yu, Mohit Bansal, and Tamara Berg. TVQA: Localized, compositional video question answering. In \emph{Proceedings of the Conference on Empirical Methods in Natural Language Processing}, 2018. URL \url{https://aclanthology.org/D18-1167/}.

\bibitem[Li et~al.(2023)]{li2023mvbench}
Kunchang Li, Yali Wang, Yinan He, et~al. MVBench: A comprehensive multimodal video understanding benchmark. \emph{arXiv preprint arXiv:2311.17005}, 2023. URL \url{https://arxiv.org/abs/2311.17005}.

\bibitem[Li et~al.(2023b)]{li2023intentqa}
Jiapeng Li, Ping Wei, Wenjuan Han, and Lifeng Fan. IntentQA: Context-aware video intent reasoning. In \emph{Proceedings of the IEEE/CVF International Conference on Computer Vision}, 2023. URL \url{https://openaccess.thecvf.com/content/ICCV2023/html/Li_IntentQA_Context-aware_Video_Intent_Reasoning_ICCV_2023_paper.html}.

\bibitem[Li et~al.(2026)]{crossvid}
Jingyao Li, Jingyun Wang, Molin Tan, Haochen Wang, Cilin Yan, Likun Shi, Jiayin Cai, Xiaolong Jiang, and Yao Hu. CrossVid: A comprehensive benchmark for evaluating cross-video reasoning of multimodal large language models. In \emph{Proceedings of the AAAI Conference on Artificial Intelligence}, pp. 6244--6252, 2026.

\bibitem[Lightman et~al.(2023)]{lightman2023letsverify}
Hunter Lightman, Vineet Kosaraju, Yura Burda, Harri Edwards, Bowen Baker, Teddy Lee, Jan Leike, John Schulman, Ilya Sutskever, and Karl Cobbe. Let's verify step by step. \emph{arXiv preprint arXiv:2305.20050}, 2023. URL \url{https://arxiv.org/abs/2305.20050}.

\bibitem[Lin et~al.(2023)]{lin2023videollava}
Bin Lin, Yang Ye, Bin Zhu, Jiaxi Cui, Munan Ning, Peng Jin, and Li Yuan. Video-LLaVA: Learning unified visual representations by alignment before projection. \emph{arXiv preprint arXiv:2311.10122}, 2023. URL \url{https://arxiv.org/abs/2311.10122}.

\bibitem[Liu et~al.(2023)]{liu2023llava}
Haotian Liu, Chunyuan Li, Qingyang Wu, and Yong Jae Lee. Visual instruction tuning. In \emph{Advances in Neural Information Processing Systems}, 2023. URL \url{https://proceedings.neurips.cc/paper_files/paper/2023/hash/6dcf277ea32ce3288914faf369fe6de0-Abstract-Conference.html}.

\bibitem[Liu et~al.(2026)]{liu2026evads}
Xianjie Liu, Yiman Hu, Liang Wu, Ping Hu, Yixiong Zou, Jian Xu, and Bo Zheng. E-VAds: An e-commerce short video understanding benchmark for MLLMs. \emph{arXiv preprint arXiv:2602.08355}, 2026. URL \url{https://arxiv.org/abs/2602.08355}.

\bibitem[Long et~al.(2025)]{long2025adsqa}
Xinwei Long, Kai Tian, Peng Xu, Guoli Jia, Jingxuan Li, Sa Yang, Yihua Shao, Kaiyan Zhang, Che Jiang, Hao Xu, Yang Liu, Jiaheng Ma, and Bowen Zhou. AdsQA: Towards advertisement video understanding. In \emph{Proceedings of the IEEE/CVF International Conference on Computer Vision}, pp. 23396--23407, 2025.

\bibitem[Maaz et~al.(2024)]{maaz2024videochatgpt}
Muhammad Maaz, Hanoona Rasheed, Salman Khan, and Fahad Khan. Video-ChatGPT: Towards detailed video understanding via large vision and language models. In \emph{Proceedings of the 62nd Annual Meeting of the Association for Computational Linguistics (Volume 1: Long Papers)}, 2024. URL \url{https://aclanthology.org/2024.acl-long.679/}.

\bibitem[Madaan et~al.(2023)]{madaan2023selfrefine}
Aman Madaan, Niket Tandon, Prakhar Gupta, et~al. Self-Refine: Iterative refinement with self-feedback. In \emph{Advances in Neural Information Processing Systems}, 2023. URL \url{https://proceedings.neurips.cc/paper_files/paper/2023/hash/91edff07232fb1b55a505a9e9f6c0ff3-Abstract-Conference.html}.

\bibitem[Mangalam et~al.(2023)]{mangalam2023egoschema}
Karttikeya Mangalam, Raiymbek Akshulakov, and Jitendra Malik. EgoSchema: A diagnostic benchmark for very long-form video language understanding. In \emph{Advances in Neural Information Processing Systems}, 2023. URL \url{https://proceedings.neurips.cc/paper_files/paper/2023/hash/90ce332aff156b910b002ce4e6880dec-Abstract-Datasets_and_Benchmarks.html}.

\bibitem[Patraucean et~al.(2023)]{patraucean2023perception}
Viorica P\u{a}tr\u{a}ucean, Lucas Smaira, Ankush Gupta, et~al. Perception Test: A diagnostic benchmark for multimodal video models. In \emph{Advances in Neural Information Processing Systems}, 2023. URL \url{https://proceedings.neurips.cc/paper_files/paper/2023/hash/8540fba4abdc7f9f7a7b1cc6cd60e409-Abstract.html}.

\bibitem[Peng et~al.(2025)]{peng2025mvueval}
Tianhao Peng, Haochen Wang, Yuanxing Zhang, Zekun Wang, Zili Wang, Gavin Chang, Jian Yang, Shihao Li, Yanghai Wang, Xintao Wang, Houyi Li, Wei Ji, Pengfei Wan, Steven Huang, Zhaoxiang Zhang, and Jiaheng Liu. MVU-Eval: Towards multi-video understanding evaluation for multimodal LLMs. In \emph{Advances in Neural Information Processing Systems}, 2025. URL \url{https://arxiv.org/abs/2511.07250}.

\bibitem[Pignatelli et~al.(2024)]{pignatelli2024credit}
Eduardo Pignatelli, Johan Ferret, Matthieu Geist, Thomas Mesnard, Hado van Hasselt, Olivier Pietquin, and Laura Toni. A survey of temporal credit assignment in deep reinforcement learning. \emph{arXiv preprint arXiv:2312.01072}, version 2, 2024. URL \url{https://arxiv.org/abs/2312.01072}.

\bibitem[Qian et~al.(2025)]{qian2025toolrl}
Cheng Qian, Emre Can Acikgoz, Qi He, Hongru Wang, Xiusi Chen, Dilek Hakkani-Tur, Gokhan Tur, and Heng Ji. ToolRL: Reward is all tool learning needs. In \emph{Advances in Neural Information Processing Systems}, 2025. URL \url{https://proceedings.neurips.cc/paper_files/paper/2025/hash/97c5b2707228e7e3fb67e4ecc2e0e607-Abstract-Conference.html}.

\bibitem[Qiu et~al.(2026)]{qiu2026agentcvr}
Yilun Qiu, Jiahe Wang, Cilin Yan, Jiayin Cai, Xiaolong Jiang, Yao Hu, and Chun Yuan. AgentCVR: Active multi-agent cross-video reasoning via script-simulated reinforcement learning. \emph{arXiv preprint arXiv:2605.29643}, 2026. URL \url{https://arxiv.org/abs/2605.29643}.

\bibitem[Radford et~al.(2022)]{radford2022whisper}
Alec Radford, Jong Wook Kim, Tao Xu, Greg Brockman, Christine McLeavey, and Ilya Sutskever. Robust speech recognition via large-scale weak supervision. \emph{arXiv preprint arXiv:2212.04356}, 2022. URL \url{https://arxiv.org/abs/2212.04356}.

\bibitem[Schick et~al.(2023)]{schick2023toolformer}
Timo Schick, Jane Dwivedi-Yu, Roberto Dess\`i, Roberta Raileanu, Maria Lomeli, Eric Hambro, Luke Zettlemoyer, Nicola Cancedda, and Thomas Scialom. Toolformer: Language models can teach themselves to use tools. In \emph{Advances in Neural Information Processing Systems}, 2023. URL \url{https://proceedings.neurips.cc/paper/2023/hash/d842425e4bf79ba039352da0f658a906-Abstract-Conference.html}.

\bibitem[Shang et~al.(2024)]{shang2024traveler}
Chuyi Shang, Amos You, Sanjay Subramanian, Trevor Darrell, and Roei Herzig. TraveLER: A modular multi-LMM agent framework for video question-answering. In \emph{Proceedings of the Conference on Empirical Methods in Natural Language Processing}, 2024. URL \url{https://aclanthology.org/2024.emnlp-main.544/}.

\bibitem[Shao et~al.(2024)]{shao2024deepseekmath}
Zhihong Shao, Peiyi Wang, Qihao Zhu, Runxin Xu, Junxiao Song, Xiao Bi, Haowei Zhang, Mingchuan Zhang, Y. K. Li, Y. Wu, and Daya Guo. DeepSeekMath: Pushing the limits of mathematical reasoning in open language models. \emph{arXiv preprint arXiv:2402.03300}, 2024. URL \url{https://arxiv.org/abs/2402.03300}.

\bibitem[Shinn et~al.(2023)]{shinn2023reflexion}
Noah Shinn, Federico Cassano, Ashwin Gopinath, Karthik Narasimhan, and Shunyu Yao. Reflexion: Language agents with verbal reinforcement learning. In \emph{Advances in Neural Information Processing Systems}, 2023. URL \url{https://proceedings.neurips.cc/paper_files/paper/2023/hash/1b44b878bb782e6954cd888628510e90-Abstract-Conference.html}.

\bibitem[Wang et~al.(2021)]{wang2021tencentavs}
Zhenzhi Wang, Liyu Wu, Zhimin Li, Jiangfeng Xiong, and Qinglin Lu. Overview of Tencent multi-modal ads video understanding challenge. \textit{arXiv preprint arXiv:2109.07951}, 2021.

\bibitem[Wang et~al.(2024a)]{wang2024videoagent}
Xiaohan Wang, Yuhui Zhang, Orr Zohar, and Serena Yeung-Levy. VideoAgent: Long-form video understanding with large language models as agents. In \emph{European Conference on Computer Vision}, pp. 58--76, 2024a.

\bibitem[Wang et~al.(2024b)]{wang2024internvideo2}
Yi Wang, Kunchang Li, Xinhao Li, et~al. InternVideo2: Scaling up foundation models for multimodal video understanding. \emph{arXiv preprint arXiv:2403.15377}, 2024b. URL \url{https://arxiv.org/abs/2403.15377}.

\bibitem[Wang et~al.(2025a)]{wang2025roll}
Weixun Wang, Shaopan Xiong, Gengru Chen, et~al. Reinforcement learning optimization for large-scale learning: An efficient and user-friendly scaling library. \emph{arXiv preprint arXiv:2506.06122}, 2025a. URL \url{https://arxiv.org/abs/2506.06122}.

\bibitem[Wang et~al.(2025b)]{wang2025videotree}
Ziyang Wang, Shoubin Yu, Elias Stengel-Eskin, Jaehong Yoon, Feng Cheng, Gedas Bertasius, and Mohit Bansal. VideoTree: Adaptive tree-based video representation for LLM reasoning on long videos. In \emph{Proceedings of the IEEE/CVF Conference on Computer Vision and Pattern Recognition}, 2025b.

\bibitem[Wu et~al.(2024)]{wu2024longvideobench}
Haoning Wu, Dongxu Li, Bei Chen, and Junnan Li. LongVideoBench: A benchmark for long-context interleaved video-language understanding. In \emph{Advances in Neural Information Processing Systems}, 2024.

\bibitem[Xiao et~al.(2021)]{xiao2021nextqa}
Junbin Xiao, Xindi Shang, Angela Yao, and Tat-Seng Chua. NExT-QA: Next phase of question-answering to explaining temporal actions. In \emph{Proceedings of the IEEE/CVF Conference on Computer Vision and Pattern Recognition}, pp. 9772--9781, 2021. URL \url{https://arxiv.org/abs/2105.08276}.

\bibitem[Xiao et~al.(2024)]{xiao2024nextgqa}
Junbin Xiao, Angela Yao, Yicong Li, and Tat-Seng Chua. Can I trust your answer? Visually grounded video question answering. In \emph{Proceedings of the IEEE/CVF Conference on Computer Vision and Pattern Recognition}, 2024. URL \url{https://openaccess.thecvf.com/content/CVPR2024/html/Xiao_Can_I_Trust_Your_Answer_Visually_Grounded_Video_Question_Answering_CVPR_2024_paper.html}.

\bibitem[Yang et~al.(2023)]{yang2023gpt4tools}
Rui Yang, Lin Song, Yanwei Li, Sijie Zhao, Yixiao Ge, Xiu Li, and Ying Shan. GPT4Tools: Teaching large language model to use tools via self-instruction. In \emph{Advances in Neural Information Processing Systems}, 2023. URL \url{https://proceedings.neurips.cc/paper_files/paper/2023/hash/e393677793767624f2821cec8bdd02f1-Abstract-Conference.html}.

\bibitem[Yao et~al.(2022)]{yao2022react}
Shunyu Yao, Jeffrey Zhao, Dian Yu, Nan Du, Izhak Shafran, Karthik Narasimhan, and Yuan Cao. ReAct: Synergizing reasoning and acting in language models. \emph{arXiv preprint arXiv:2210.03629}, 2022. URL \url{https://arxiv.org/abs/2210.03629}.

\bibitem[Yu et~al.(2019)]{yu2019activitynetqa}
Zhou Yu, Dejing Xu, Jun Yu, Ting Yu, Zhou Zhao, Yueting Zhuang, and Dacheng Tao. ActivityNet-QA: A dataset for understanding complex web videos via question answering. In \emph{Proceedings of the AAAI Conference on Artificial Intelligence}, volume 33, pp. 9127--9134, 2019. doi: 10.1609/aaai.v33i01.33019127.

\bibitem[Zhang et~al.(2024)]{zhang2024llavavideo}
Yuanhan Zhang, Jinming Wu, Wei Li, Bo Li, Zejun Ma, Ziwei Liu, and Chunyuan Li. LLaVA-Video: Video instruction tuning with synthetic data. \emph{arXiv preprint arXiv:2410.02713}, 2024. URL \url{https://arxiv.org/abs/2410.02713}.

\bibitem[Yu et~al.(2023)]{yu2023sevila}
Shoubin Yu, Jaemin Cho, Prateek Yadav, and Mohit Bansal. Self-chained image-language model for video localization and question answering. In \emph{Advances in Neural Information Processing Systems}, 2023. URL \url{https://proceedings.neurips.cc/paper_files/paper/2023/hash/f22a9af8dbb348952b08bd58d4734b50-Abstract-Conference.html}.

\bibitem[Zelikman et~al.(2022)]{zelikman2022star}
Eric Zelikman, Yuhuai Wu, Jesse Mu, and Noah Goodman. STaR: Bootstrapping reasoning with reasoning. In \emph{Advances in Neural Information Processing Systems}, 2022. URL \url{https://papers.nips.cc/paper/2022/hash/639a9a172c044fbb64175b5fad42e9a5-Abstract-Conference.html}.

\bibitem[Zhang et~al.(2024)]{zhang2024llovi}
Ce Zhang, Taixi Lu, Md Mohaiminul Islam, Ziyang Wang, Shoubin Yu, Mohit Bansal, and Gedas Bertasius. A simple LLM framework for long-range video question-answering. In \emph{Proceedings of the Conference on Empirical Methods in Natural Language Processing}, 2024. URL \url{https://aclanthology.org/2024.emnlp-main.1209/}.

\bibitem[Zhang et~al.(2025)]{zhang2025videoads}
Zheyuan Zhang, Wanying Dou, Linkai Peng, Hongyi Pan, Ulas Bagci, and Boqing Gong. VideoAds for fast-paced video understanding. In \emph{Proceedings of the IEEE/CVF International Conference on Computer Vision}, pp. 21812--21821, 2025.

\bibitem[Zhang et~al.(2026)]{zhang2026eva}
Yaolun Zhang, Ruohui Wang, Jiahao Wang, Yepeng Tang, Xuanyu Zheng, Haonan Duan, Hao Lu, Hanming Deng, and Lewei Lu. EVA: Efficient reinforcement learning for end-to-end video agent. In \emph{Proceedings of the IEEE/CVF Conference on Computer Vision and Pattern Recognition}, 2026. URL \url{https://arxiv.org/abs/2603.22918}.

\bibitem[Zheng et~al.(2024a)]{zheng2024processbench}
Chujie Zheng, Zhenru Zhang, Beichen Zhang, Runji Lin, Keming Lu, Bowen Yu, Dayiheng Liu, Jingren Zhou, and Junyang Lin. ProcessBench: Identifying process errors in mathematical reasoning. \emph{arXiv preprint arXiv:2412.06559}, 2024a. URL \url{https://arxiv.org/abs/2412.06559}.

\bibitem[Zheng et~al.(2024b)]{zheng2024llamafactory}
Yaowei Zheng, Richong Zhang, Junhao Zhang, Yanhan Ye, and Zheyan Luo. LlamaFactory: Unified efficient fine-tuning of 100+ language models. In \emph{Proceedings of the 62nd Annual Meeting of the Association for Computational Linguistics (Volume 3: System Demonstrations)}, pp. 400--410. Association for Computational Linguistics, 2024b. doi: 10.18653/v1/2024.acl-demos.38. URL \url{https://aclanthology.org/2024.acl-demos.38/}.

\bibitem[Zhou et~al.(2024)]{zhou2024mlvu}
Junjie Zhou, Yan Shu, Bo Zhao, et~al. MLVU: Benchmarking multitask long video understanding. \emph{arXiv preprint arXiv:2406.04264}, 2024. URL \url{https://arxiv.org/abs/2406.04264}.

\bibitem[Zhu et~al.(2025)]{zhu2025cvbench}
Nannan Zhu, Yonghao Dong, Teng Wang, et~al. CVBench: Benchmarking cross-video collaboration for complex multimodal reasoning. \emph{arXiv preprint arXiv:2508.19542}, 2025. URL \url{https://arxiv.org/abs/2508.19542}.

\end{thebibliography}

\clearpage
\appendix
\section{Annotation Guidelines}
AdsCVR annotations compare two to four e-commerce videos along six dimensions: TS, CS, ID, VC, IC, and OD. Data construction first applies category-balanced quality filtering and timestamped ASR/OCR extraction, then uses Gemini 3.5 Flash to generate structured descriptions and salient tags. Candidate questions are generated as $(\text{question},\allowbreak \text{options},\allowbreak \text{answer},\allowbreak \text{clues})$ tuples. A hierarchical verifier uses Qwen3-VL-8B for prompt-only shortcut and difficulty checks, followed by GPT-5.5 checks for answer completeness, multi-video dependence, logical validity, and multimodal evidence requirements.

\subsection{Qualitative Examples Across Dimensions}
\begin{figure*}[!htbp]
  \centering
  \includegraphics[width=0.495\textwidth]{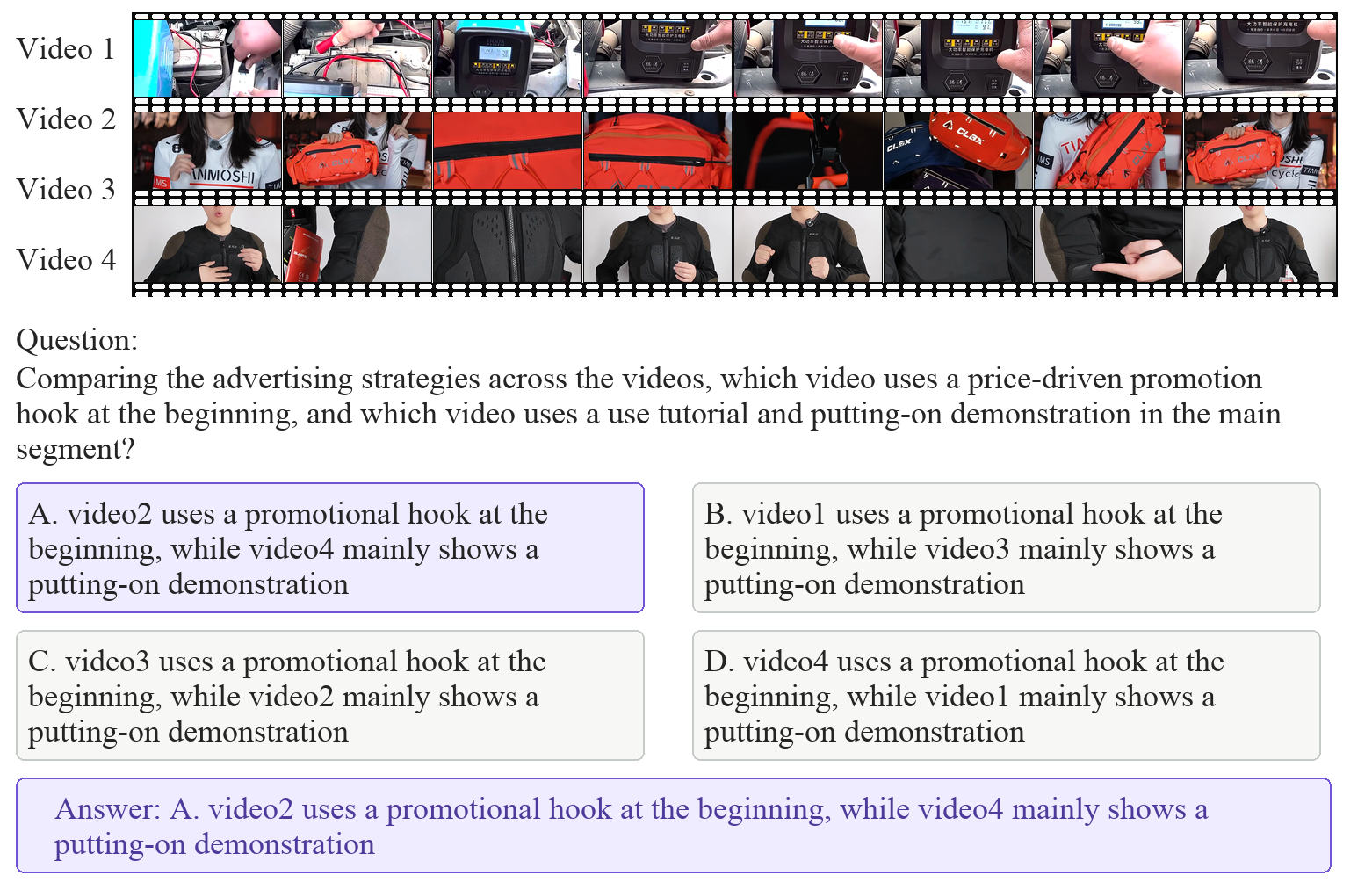}\hfill
  \includegraphics[width=0.495\textwidth]{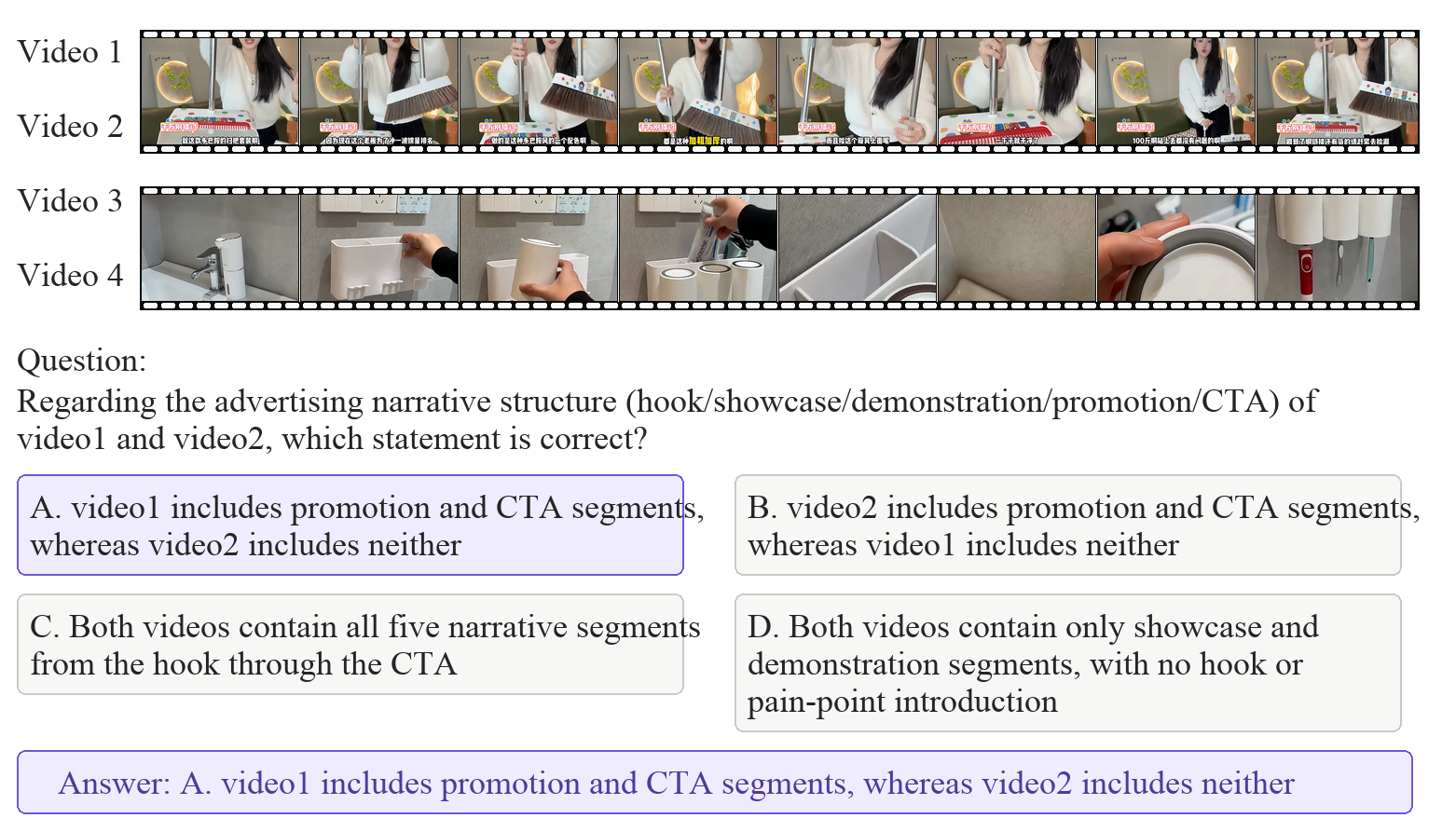}
  \caption{Temporal Structure (TS) examples comparing how advertisements organize events and product claims.}
  \label{fig:adscvr_cases_ts}
\end{figure*}

\begin{figure*}[!htbp]
  \centering
  \includegraphics[width=0.495\textwidth]{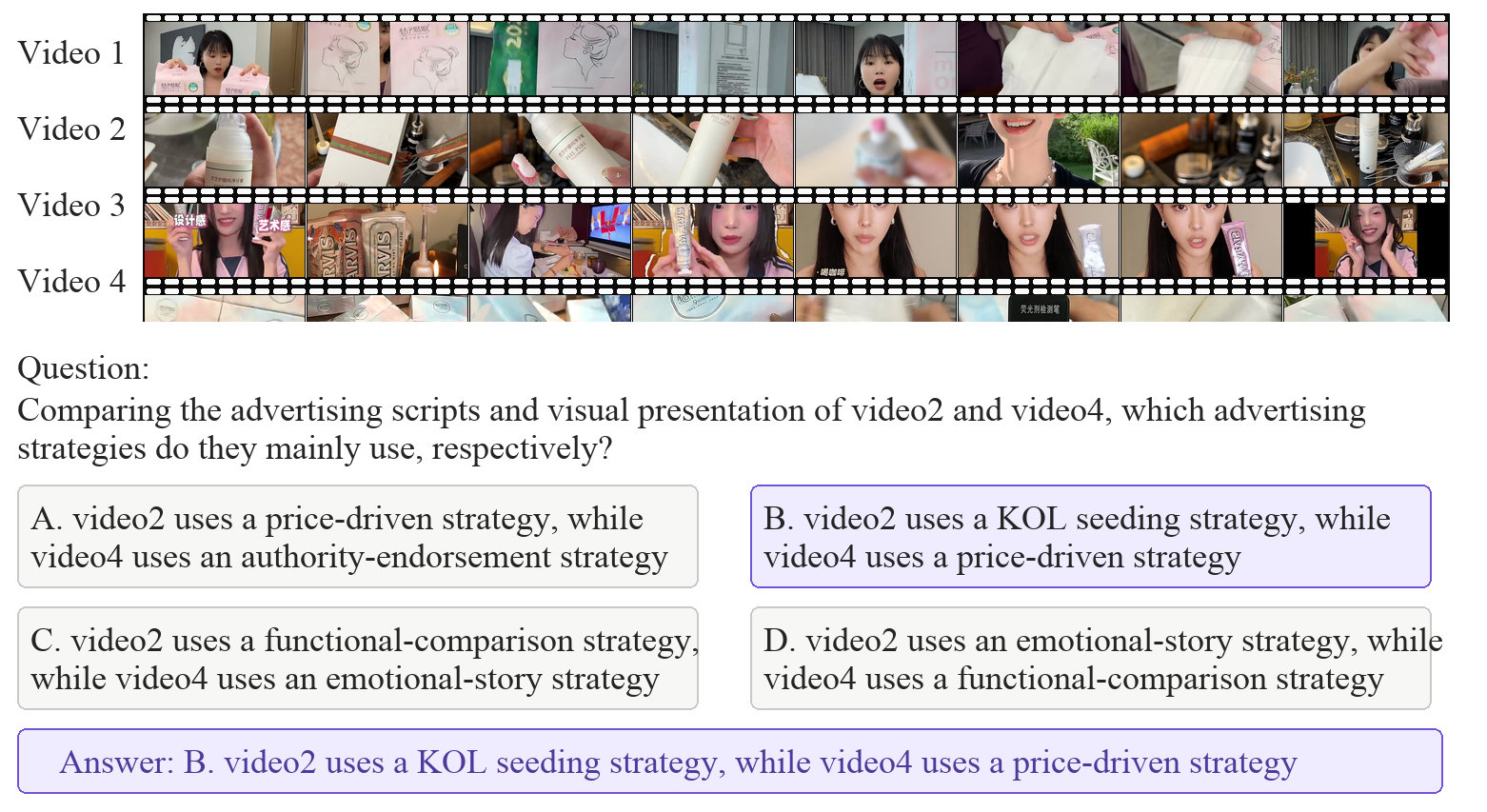}\hfill
  \includegraphics[width=0.495\textwidth]{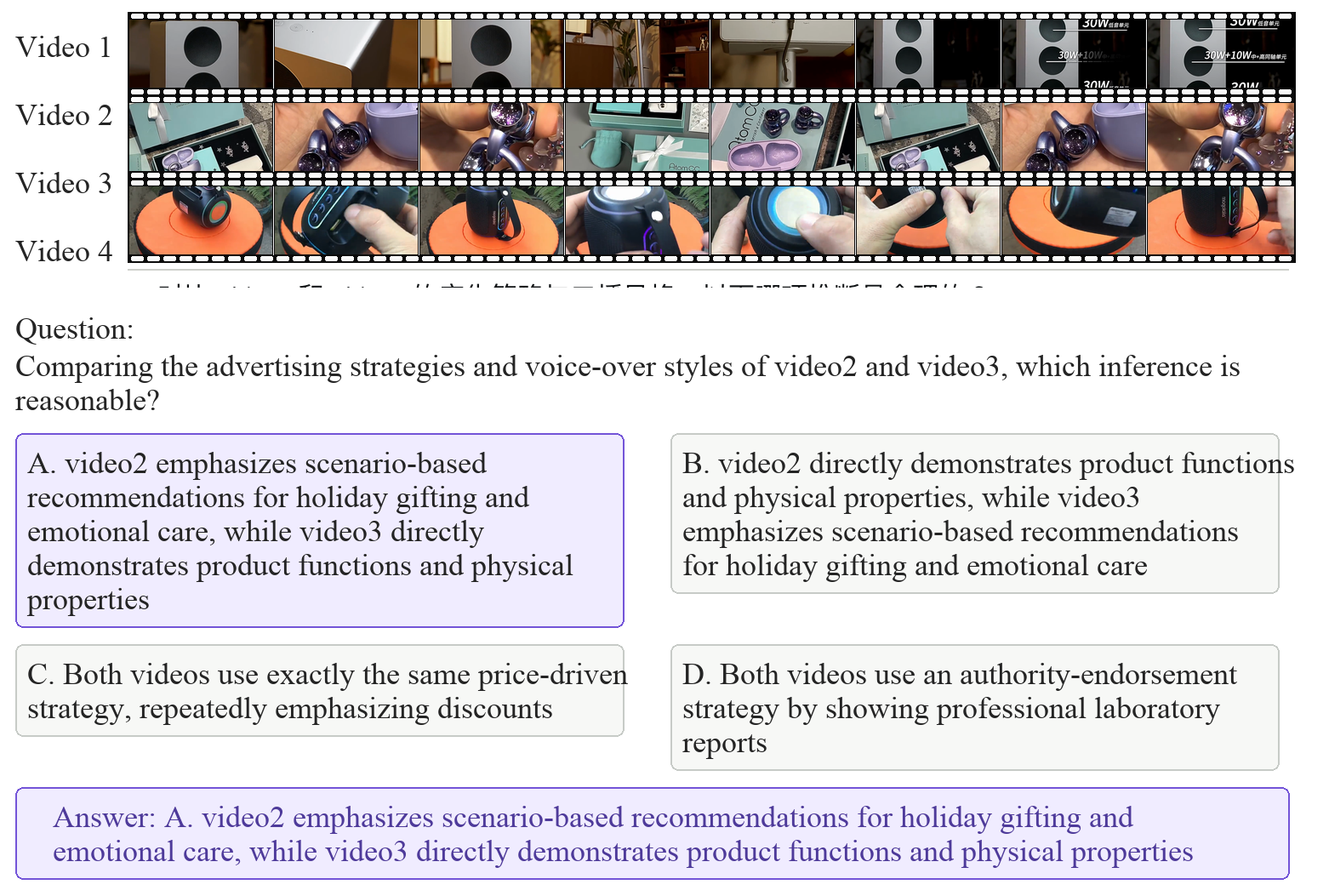}
  \caption{Creative Strategy (CS) examples comparing how advertisements attract attention and communicate product value.}
  \label{fig:adscvr_cases_cs}
\end{figure*}

\begin{figure*}[!htbp]
  \centering
  \includegraphics[width=0.495\textwidth]{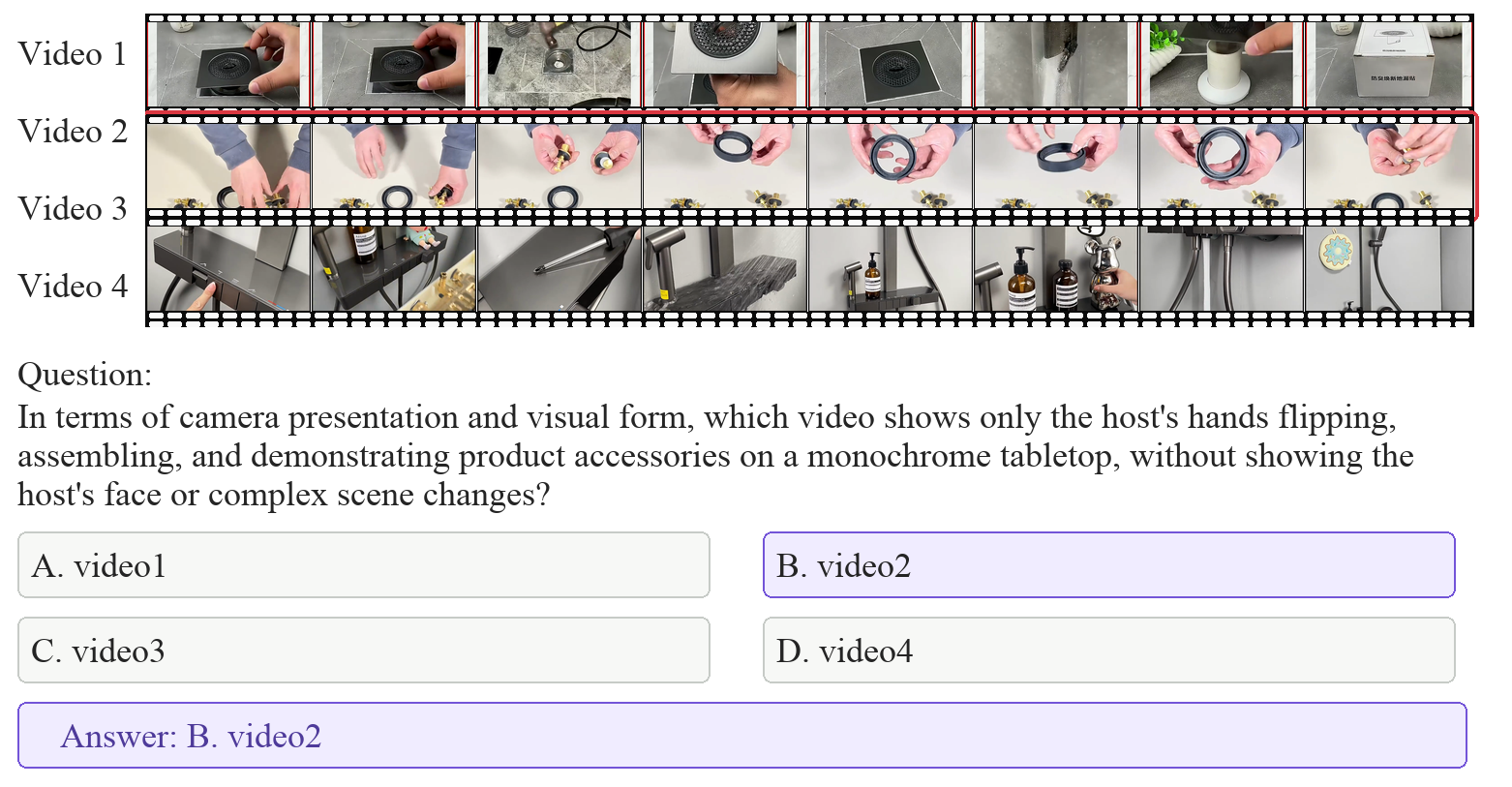}\hfill
  \includegraphics[width=0.495\textwidth]{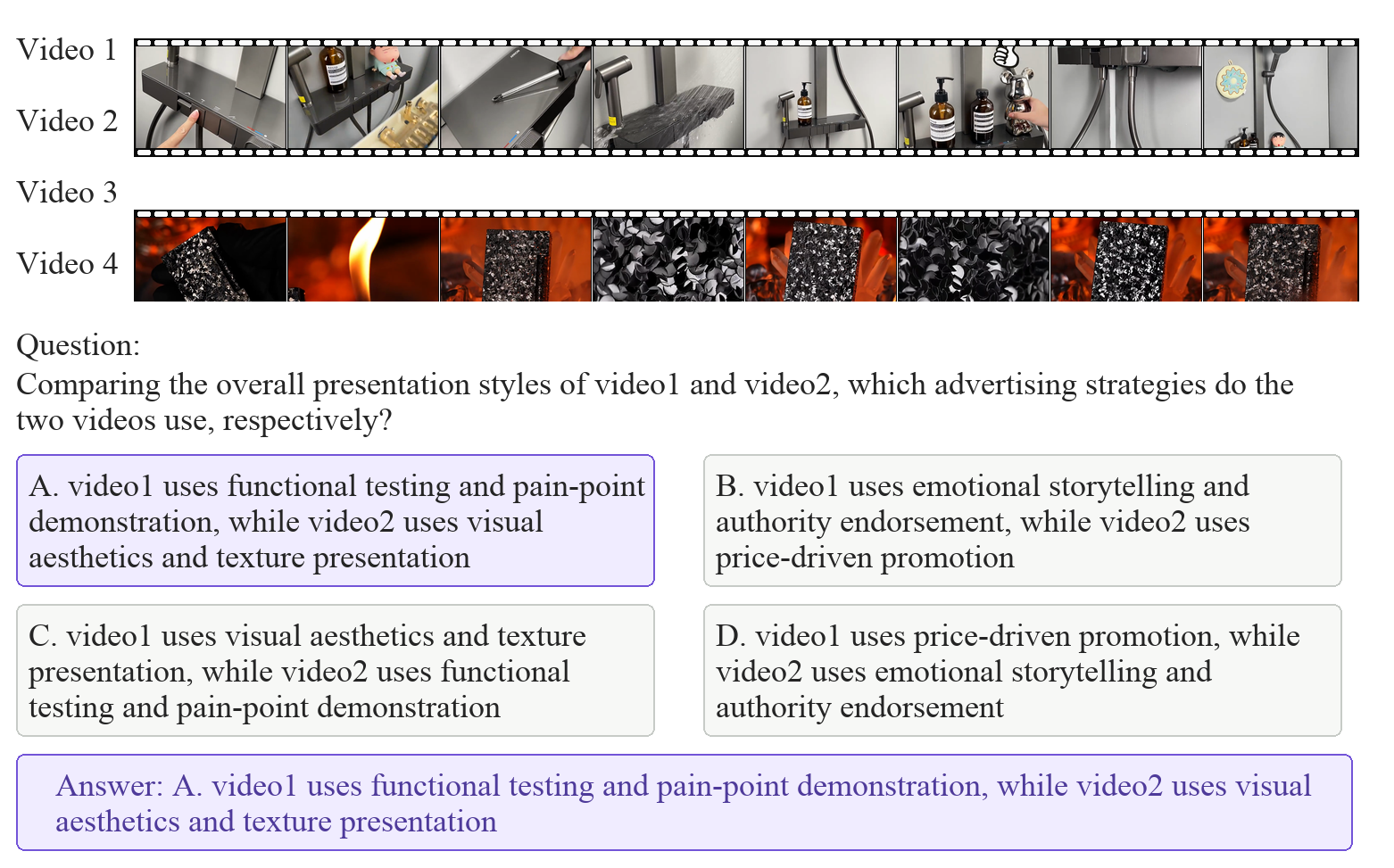}
  \caption{Outlier Detection (OD) examples comparing advertisements across product categories.}
  \label{fig:adscvr_cases_od}
\end{figure*}

\begin{figure*}[!htbp]
  \centering
  \includegraphics[width=0.495\textwidth]{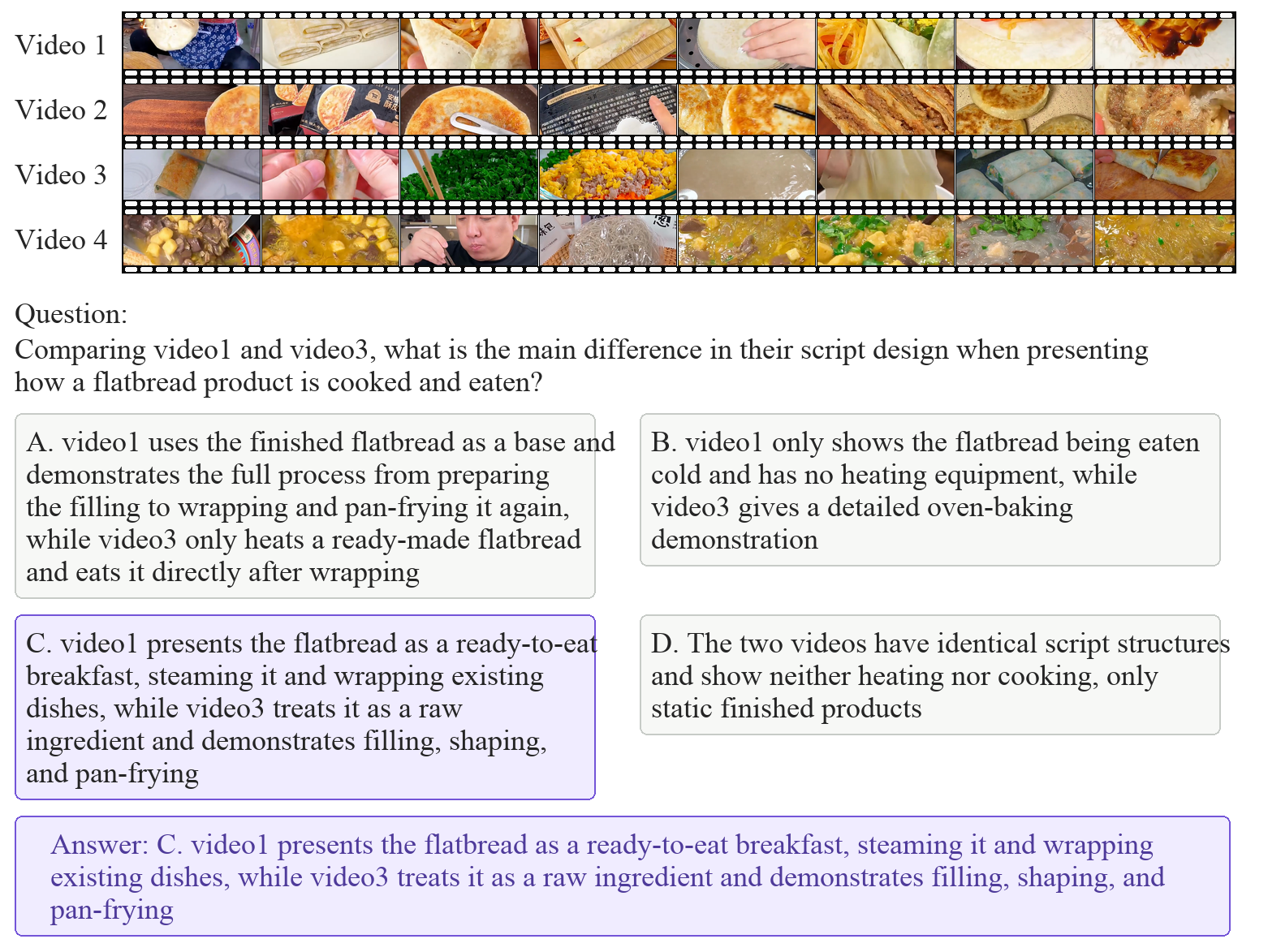}\hfill
  \includegraphics[width=0.495\textwidth]{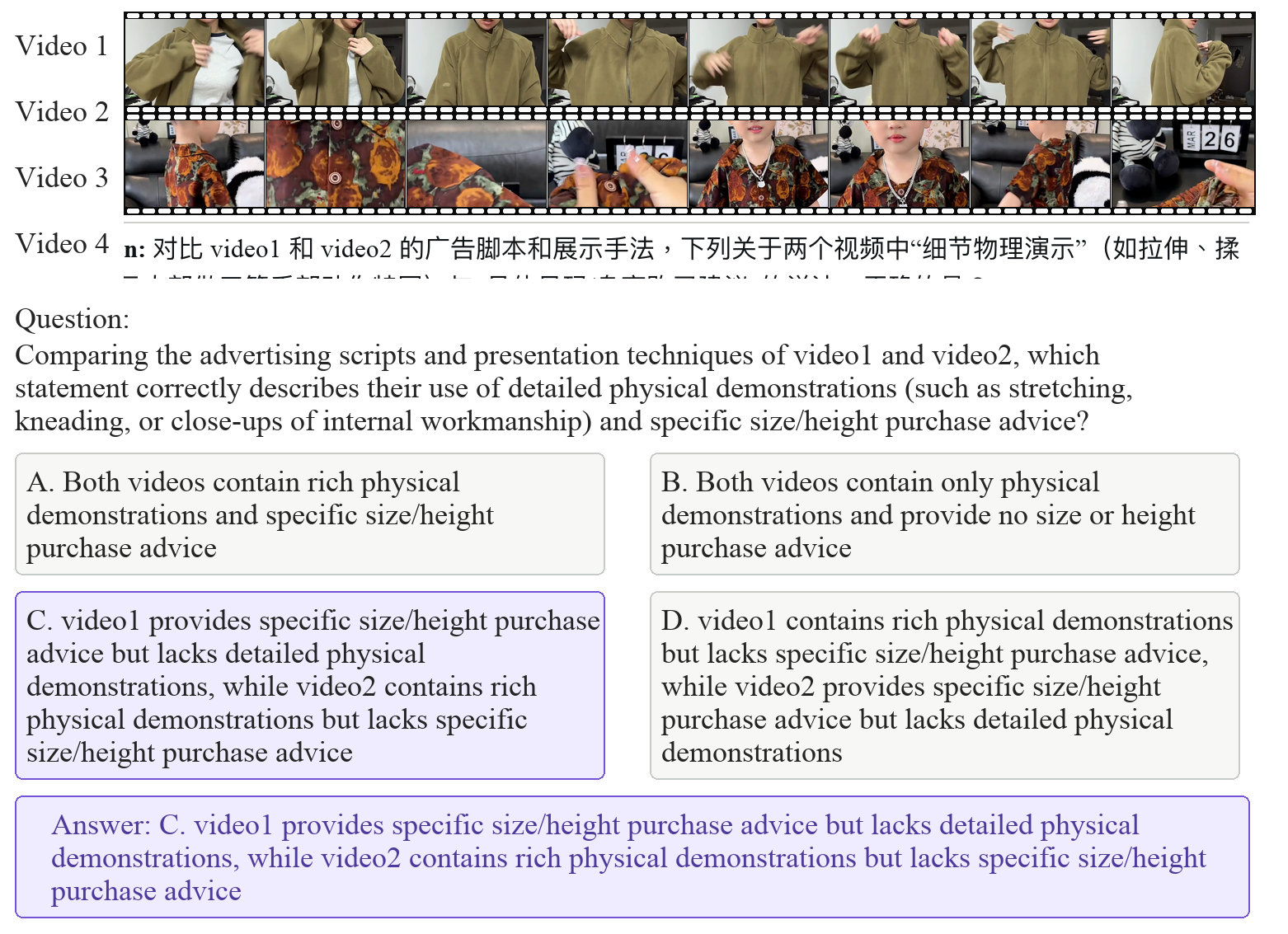}
  \caption{Intra-class Comparison (IC) examples comparing substantive similarities and differences among advertisements for related products.}
  \label{fig:adscvr_cases_ic}
\end{figure*}

\begin{figure*}[!htbp]
  \centering
  \includegraphics[width=0.495\textwidth]{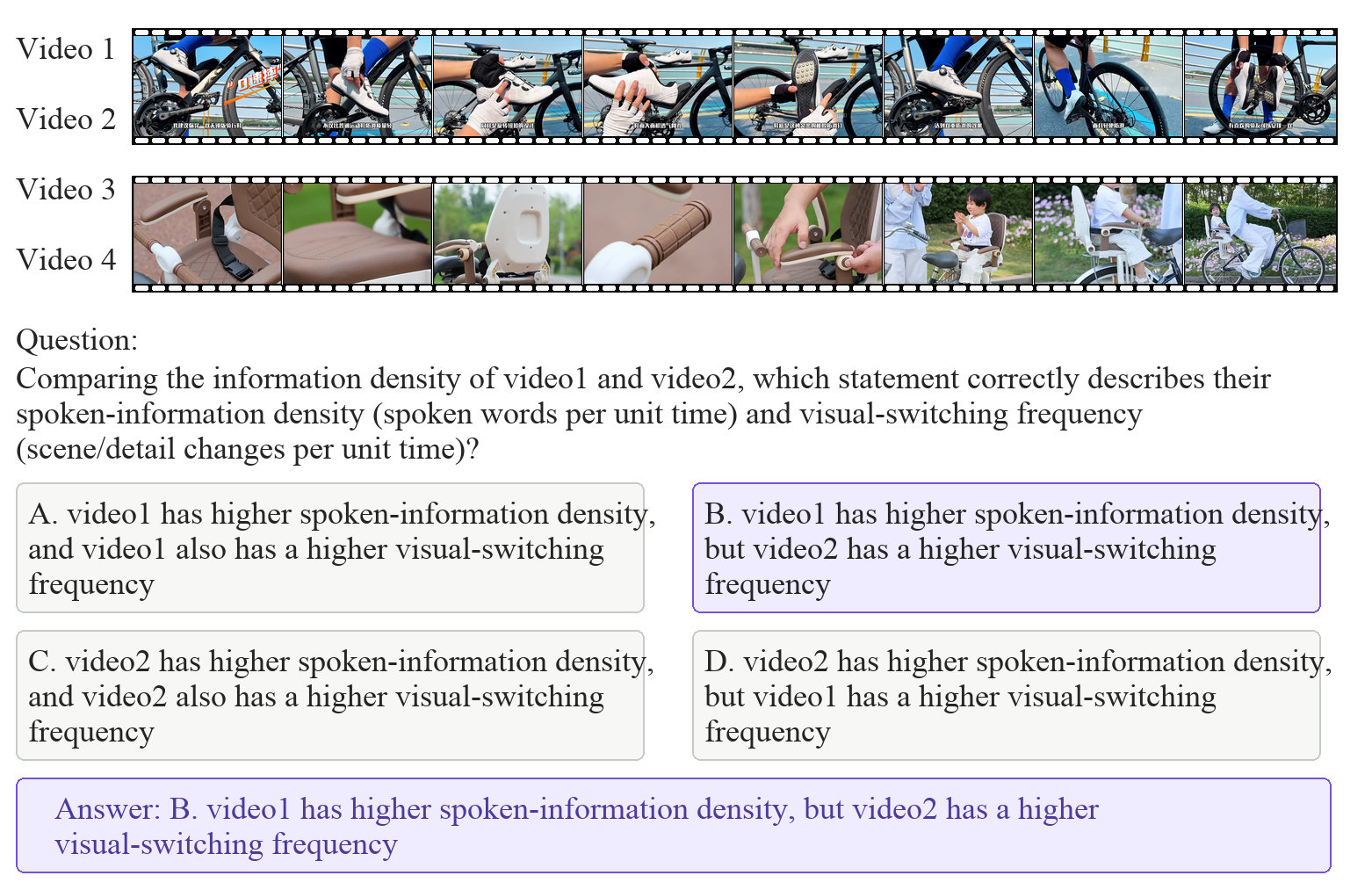}\hfill
  \includegraphics[width=0.495\textwidth]{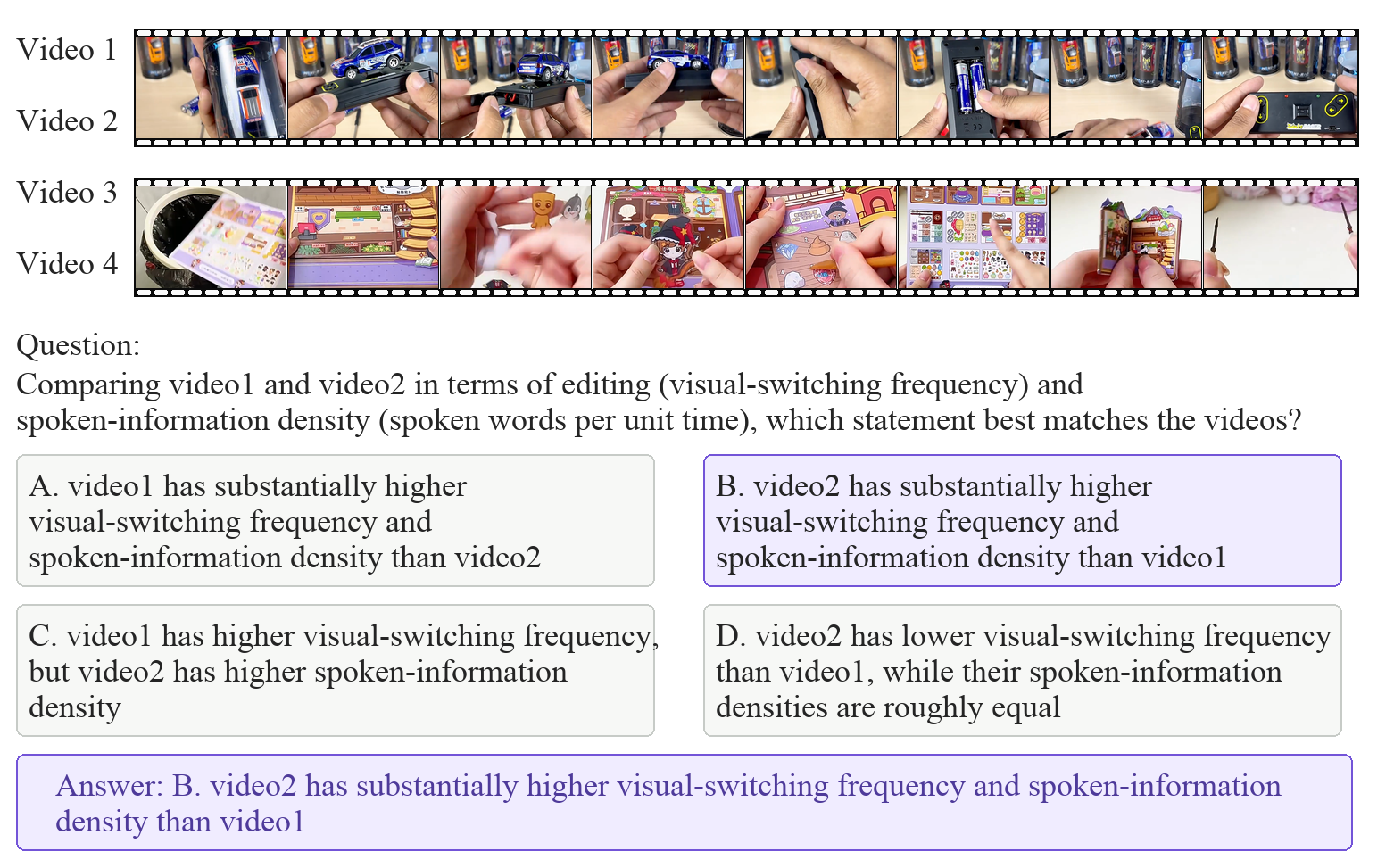}
  \caption{Information Density (ID) examples comparing the amount and pace of product-related spoken and visual information.}
  \label{fig:adscvr_cases_id}
\end{figure*}

\begin{figure*}[!htbp]
  \centering
  \includegraphics[width=0.495\textwidth]{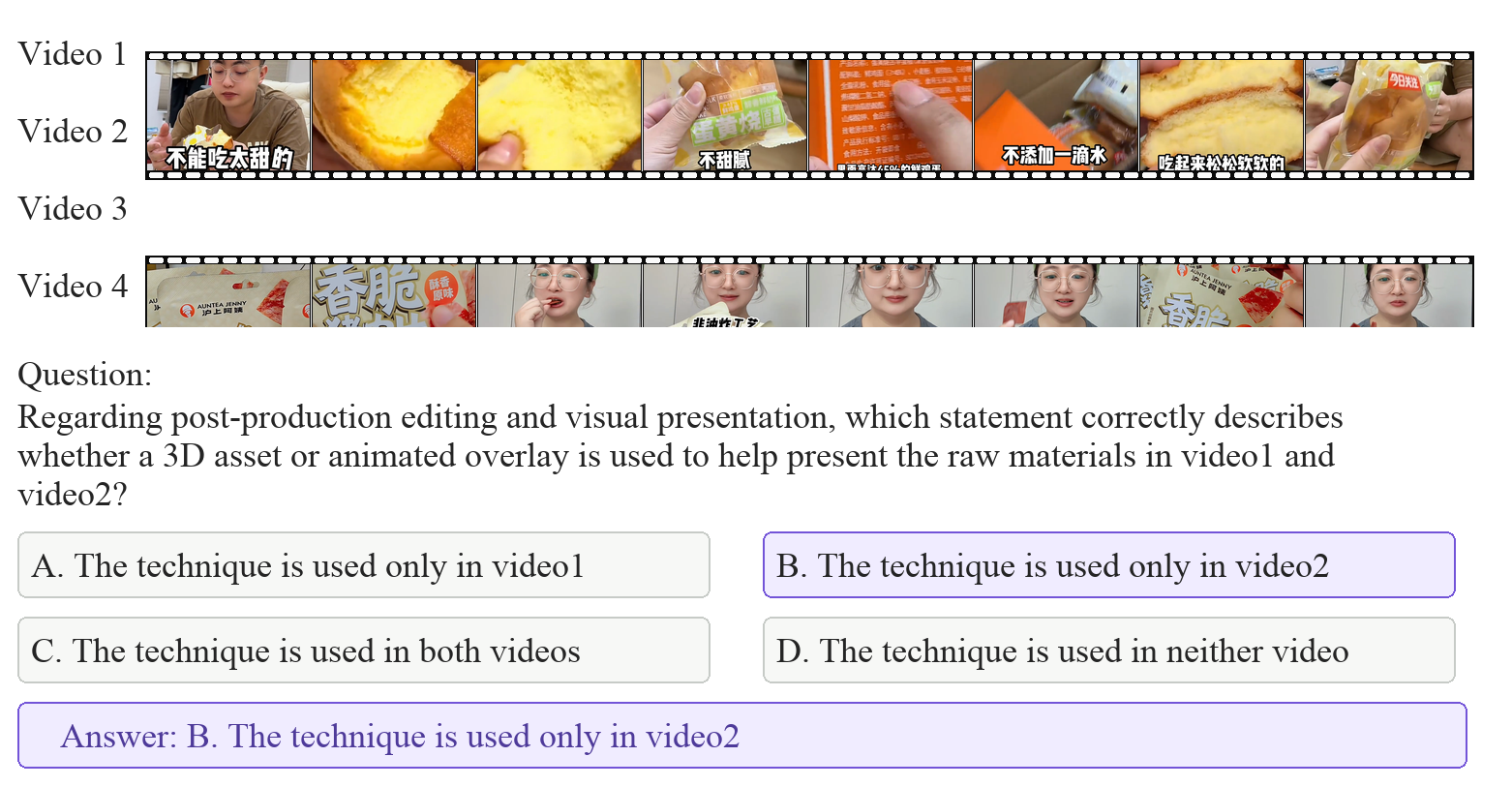}\hfill
  \includegraphics[width=0.495\textwidth]{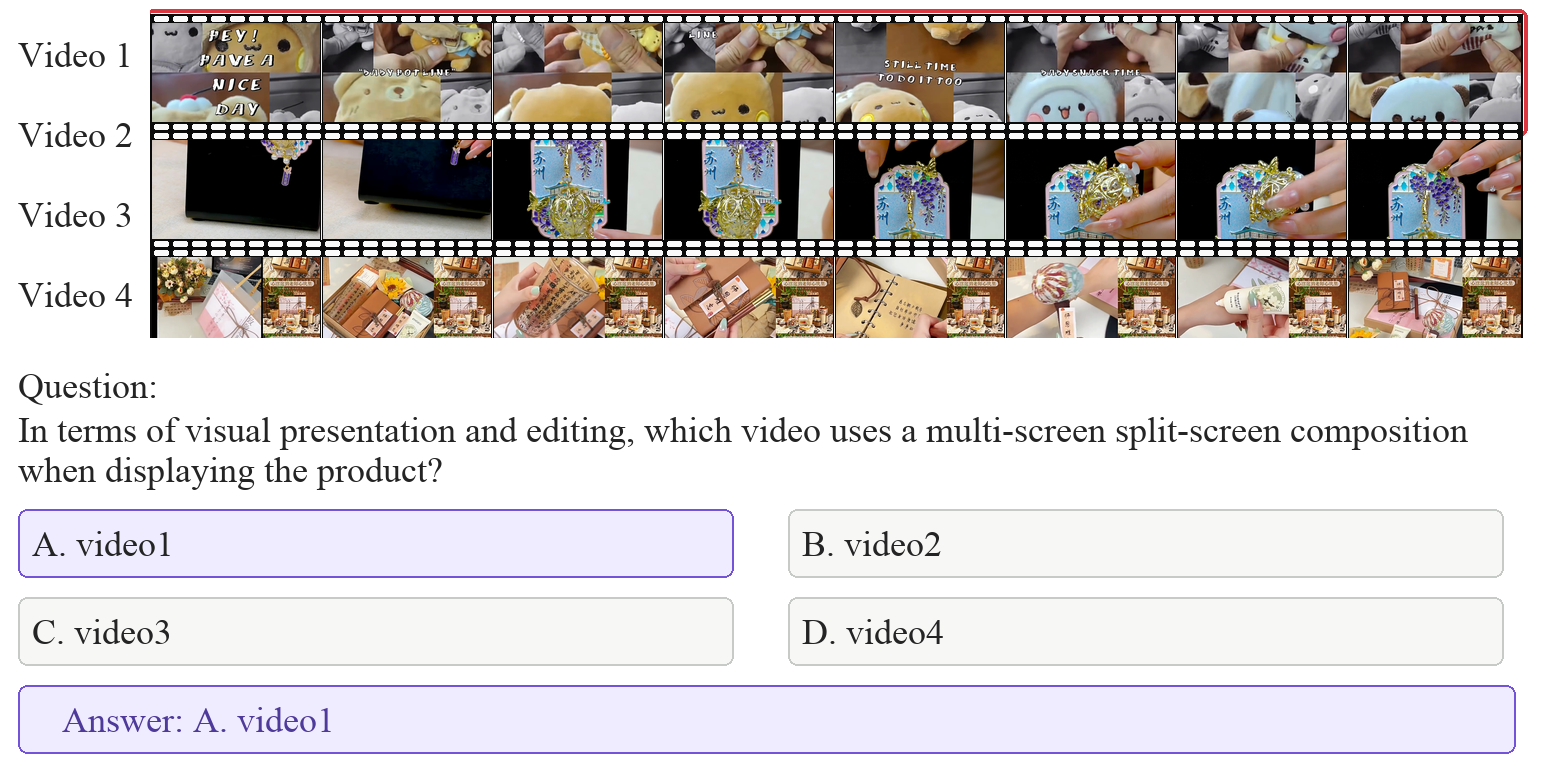}
  \caption{Visual Composition (VC) examples comparing framing, editing, transitions, and the functional role of individual shots.}
  \label{fig:adscvr_cases_vc}
\end{figure*}

Figures~\ref{fig:adscvr_cases_ts}--\ref{fig:adscvr_cases_vc} show selected cases for each of the six dimensions.
\FloatBarrier

\clearpage
\subsection{Multi-view Annotation Prompt}
\label{app:multiview_prompt}
The following prompt is used to generate the four-view annotations described in Section~\ref{sec:qa_generation}. The placeholders in the user prompt are filled with the corresponding metadata for each video.

\paragraph{System prompt.}
{\scriptsize
\begin{verbatim}
SYSTEM_PROMPT_ALL = (
    "You are a multi-role video synthesis analyst. Analyze the advertising video
     in depth by acting as each of the following four roles.\n\n"

    "[General requirements (strictly followed by all roles)]\n"
    "- Ground every fact: describe only content clearly visible in the video or
     reasonably inferable; do not invent dialogue, background stories, or
     off-screen elements\n"
    "- Segmentation: prioritize the [shot time segmentation] in the auxiliary
     information. If it is unavailable, segment by action changes, shot
     transitions, or scene changes; uniform time segmentation is prohibited\n"
    "- Caption length for each segment: 20-50 words\n\n"

    "[Role 1: Event analyst (temporal-event view)]\n"
    "Describe the main events occurring in each shot:\n"
    "  Temporal dynamics: specific actions, motion trajectories, human
     interactions, object manipulation, and state changes\n"
    "  Spatial context: scene environment and positions of key objects; describe
     them only when they change\n\n"

    "[Role 2: Scene analyst (object-and-space view)]\n"
    "Describe the people, objects, and scene in each shot:\n"
    "  Spatial details: setting, spatial relations among objects (left/right,
     above/below, front/behind, near/far), people's positions and postures,
     color and material properties, and composition\n"
    "  Key objects: objects relevant to the video topic or the current frame\n\n"

    "[Role 3: Creative analyst (cinematographic view)]\n"
    "Analyze the filming and narrative techniques of each shot:\n"
    "  Cinematic language: camera movement (push/pull, pan, follow, or static),
     shot scale (close-up, medium, or wide), and camera angle (high, low, or
     eye-level)\n"
    "  Narrative structure: the function of the shot in the overall script
     (introduction, demonstration, conflict, transition, or ending)\n"
    "  Emotional tone: observable emotional style and pacing characteristics\n\n"

    "[Role 4: Commercial analyst (holistic view, no segmentation)]\n"
    "Make an overall commercial judgment about the video:\n"
    "  audience: target audience, formatted as [price positioning + group], no
     more than 20 words and without a qualifier prefix\n"
    "    Correct example: [mid-to-low-priced consumer goods, students and
     ordinary office workers]\n"
    "    Incorrect example: [consumers who value the user experience and dislike
     a sticky feeling]\n"
    "  ad_logic: core marketing logic, explaining in one sentence how the
     advertisement drives the purchase decision\n\n"

    "Output 3-5 additional tags for each dimension: short phrases, each no more
     than 10 words, representing the core category or key point of that
     dimension.\n"
    "Output all four dimensions at once, strictly in JSON format, with no content
     outside the JSON."
)
\end{verbatim}
}

\paragraph{User prompt.}
{\scriptsize
\begin{verbatim}
USER_PROMPT_ALL = """The following auxiliary information is provided for the video:

ASR transcript: {asr}
Product title: {item_title}
Video title: {content_title}
Merchant name: {shop_name}
Video description: {summary}
Shot time segmentation: {shot_detect}
Product category: {item_cate_level1_name} / {item_cate_level1_name2}

Watch the video frame by frame. Strictly follow the role requirements in the
system instruction and use shot time segmentation as the basis for segmentation.
If shot time segmentation is unavailable, use action changes, shot transitions,
or scene changes. Output the analysis for all four dimensions at once.

Output format (strict JSON, with no other content):
{{"events":{{"segments":[{{"start":0,"end":5,"caption":"Person A walks to the display,
 picks up the product, and the camera follows the hand movement"}}],"tags":["product
 demonstration","human interaction"]}},"objects":{{"segments":[{{"start":0,"end":5,
"caption":"A white display is centered, the product is on the left side of the
 surface, and a person stands in the right foreground"}}],"tags":["indoor setting",
"product close-up"]}},"creative":{{"segments":[{{"start":0,"end":5,"caption":"A
 static medium shot from an eye-level angle introduces the product"}}],"tags":["static
 camera","product introduction"]}},"commercial":{{"audience":"Mid-to-low-priced consumer
 goods for students and ordinary office workers","ad_logic":"A realistic usage scenario
 lowers the decision barrier and encourages impulse purchase","tags":["mass-market
 consumption","scenario-based promotion"]}}}}
"""
\end{verbatim}
}

\subsection{Dataset Split and Video-Overlap Audit}
\label{app:dataset_split}
AdsCVR is split randomly into training and test partitions. To obtain a stricter test subset for cross-video transfer, we separately extracted test examples and performed a video-overlap audit against the training partition before evaluation. The resulting strictly non-overlapping subset contains ~1k samples. Table~\ref{tab:split_overlap_audit} aligns the audited test-subset accuracy with the mixed-source AdSeek result across the six dimensions. The high accuracy retained by the audited subset, together with the explicit overlap review, supports the robustness of our method to cross-video transfer.

\begin{table}[H]
\centering
\caption{Accuracy (\%) on the strictly audited test subset and with mixed-source AdSeek across the six AdsCVR dimensions. AdsCVR is randomly split, and the separately extracted test subset is audited for video overlap with the training partition before evaluation.}
\label{tab:split_overlap_audit}
\small
\setlength{\tabcolsep}{3pt}
\renewcommand{\arraystretch}{1.05}
\MDTableFit{\begin{tabular}{@{}lrrrrrrr@{}}
\toprule
Variant & TS & CS & ID & IC & OD & VC & Average \\
\midrule
Strictly audited test subset & 86.7 & 78.5 & 78.5 & 75.6 & 74.6 & 78.5 & 78.73 \\
AdSeek (mixed) & 77.0 & 73.4 & 70.8 & 75.6 & 74.8 & 74.2 & 74.30 \\
\bottomrule
\end{tabular}}
\end{table}
\FloatBarrier

\section{Implementation Details}
\label{app:implementation}
Table~\ref{tab:sft_hparams} summarizes the SFT hyperparameters used for AdsCVR, and Table~\ref{tab:rl_hparams} reports the corresponding RL configuration. The default SFT launch uses 16 PPUs; the RL run uses 32 PPUs.

\FloatBarrier
\begin{table}[H]
\centering
\caption{SFT hyperparameters for AdsCVR.}
\label{tab:sft_hparams}
\small
\begin{tabularx}{\linewidth}{@{}>{\raggedright\arraybackslash}p{0.42\linewidth}>{\raggedright\arraybackslash}X@{}}
\toprule
Hyperparameter & Value \\
\midrule
Base model & Qwen3-VL-8B-Instruct \\
Fine-tuning & Full-parameter SFT; FlashAttention 2; Liger kernel \\
Template & \texttt{qwen3\_vl\_nothink} \\
Cutoff length & 51,200 tokens \\
Per-device batch size & 1 \\
Gradient accumulation & 2 \\
Learning rate & $1\times10^{-5}$ \\
Scheduler / warmup & Cosine / warmup ratio 0.1 \\
Epochs & 5 \\
Precision / memory & BF16; gradient checkpointing \\
Checkpointing & Save every 10 steps; keep all checkpoints \\
Distributed training & 16 PPUs \\
\bottomrule
\end{tabularx}
\end{table}

\begin{table}[H]
\centering
\caption{RL hyperparameters for AdsCVR.}
\label{tab:rl_hparams}
\small
\begin{tabularx}{\linewidth}{@{}>{\raggedright\arraybackslash}p{0.42\linewidth}>{\raggedright\arraybackslash}X@{}}
\toprule
Hyperparameter & Value \\
\midrule
Base model & Qwen3-VL-8B-Instruct \\
Training steps / checkpoint interval & 500 / 20 steps \\
Rollout / validation batch size & 512 / 256 \\
Learning rate / scheduler & $5\times10^{-7}$ / constant \\
Weight decay / warmup & $1\times10^{-2}$ / 50 steps \\
Per-device batch / accumulation & 1 / 16 \\
Advantage estimator & GRPO; no advantage whitening \\
PPO epochs & 1 \\
KL regularization & KL loss coefficient $10^{-2}$; initial KL coefficient 0.001 \\
Entropy coefficient & 0.01 \\
Reward weights & Accuracy 1.0; format 0.1 \\
Rollout groups & group size 8 \\
Filtering / retries & $\texttt{MAX\_TRY}=3$ \\
Precision  & BF16 \\
Distributed training & 32 PPUs \\
\bottomrule
\end{tabularx}
\end{table}
\FloatBarrier

\section{Video Complexity Metrics}
\label{app:video_complexity}
\label{app:jpeg_complexity}
We detail the temporal and spatial complexity proxies used for video filtering in Section~\ref{sec:video_filtering}.

\paragraph{Frame sampling and temporal complexity.}
We sample frames at approximately 1 fps and uniformly retain at most 16 frames per video. For the $T$ selected frames, let
$\mathbf{u}_t=\mathbf{z}_t/(\|\mathbf{z}_t\|_2+\delta)$
be the normalized DINOv3 feature, with $\delta=10^{-12}$.
We compute:
\begin{equation}
\begin{aligned}
c_t &= 1-
\frac{\sum_{j\in\mathcal{N}_t}w_{tj}\mathbf{u}_t^\top\mathbf{u}_j}
{\sum_{j\in\mathcal{N}_t}w_{tj}+\delta},\\
\mathrm{Dcpx}(V) &= \sum_{t=1}^{T}c_t,\qquad
\mathrm{Vcpx}(V)=\frac{100}{T}\mathrm{Dcpx}(V),
\end{aligned}
\label{eq:video_complexity}
\end{equation}
where $\mathcal{N}_t=\{j:1\leq j\leq T,\ 0<|j-t|\leq d\}$,
$w_{tj}=\exp(-|j-t|/(2d))$, and $d=8$.
Both metrics are set to zero for $T\leq1$.
Higher $\mathrm{Dcpx}$ and $\mathrm{Vcpx}$ indicate greater cumulative and average visual variation, respectively.

\paragraph{Spatial complexity.}
For the same $T$ selected frames $I_1,\ldots,I_T$, we estimate spatial complexity using the average JPEG-encoded bytes per pixel:
\begin{equation}
\mathrm{Jcpx}(V)=\frac{1}{T}\sum_{t=1}^{T}
\frac{B_{\mathrm{JPEG}}(I_t)}{H_tW_t},
\label{eq:jpeg_complexity}
\end{equation}
where $B_{\mathrm{JPEG}}(I_t)$ is the encoded size in bytes and $H_t\times W_t$ is the frame resolution. We use quality 95, 4:2:0 chroma subsampling, and optimized coding. Higher $\mathrm{Jcpx}$ indicates lower spatial compressibility under these fixed encoding settings.

\end{document}